\documentclass[lettersize,journal]{IEEEtran}
\usepackage{amsmath,amsfonts}
\usepackage{array}
\usepackage[caption=false,font=normalsize,labelfont=sf,textfont=sf]{subfig}
\usepackage{textcomp}
\usepackage{stfloats}
\usepackage{url}
\usepackage{verbatim}
\usepackage{graphicx}
\usepackage{cite}
\usepackage{booktabs}
\usepackage{pifont}
\usepackage{enumitem}
\usepackage{amssymb}
\usepackage{multirow}
\usepackage{algorithm}
\usepackage{algorithmicx}
\usepackage{algpseudocode}
\usepackage{soul}
\usepackage{xcolor}
\definecolor{revcolor}{RGB}{255,255,180}
\sethlcolor{revcolor}
\newcommand{\rev}[1]{#1}

\begin{document}

\title{\rev{EmoSpace: Immersive Affective Image Generation Guided by Fine-Grained Emotion Prototypes}}

\author{Bingyuan~Wang, Xingbei~Chen, Zongyang~Qiu, Lin-Ping~Yuan, and~Zeyu~Wang
\thanks{Corresponding author: Zeyu Wang.}
\thanks{Bingyuan~Wang, Xingbei~Chen, and Zongyang~Qiu are with The Hong Kong University of Science and Technology (Guangzhou) (e-mail: bwang667@connect.hkust-gz.edu.cn; xchen053@connect.hkust-gz.edu.cn; zane.zy.qiu@gmail.com).}%
\thanks{Lin-Ping~Yuan is with The Hong Kong University of Science and Technology (e-mail: yuanlp@ust.hk).}
\thanks{Zeyu~Wang is with The Hong Kong University of Science and Technology (Guangzhou) and The Hong Kong University of Science and Technology (e-mail: zeyuwang@ust.hk).}}



\maketitle

\begin{figure*}[!t]
\rev{
\centering
\includegraphics[width=\linewidth]{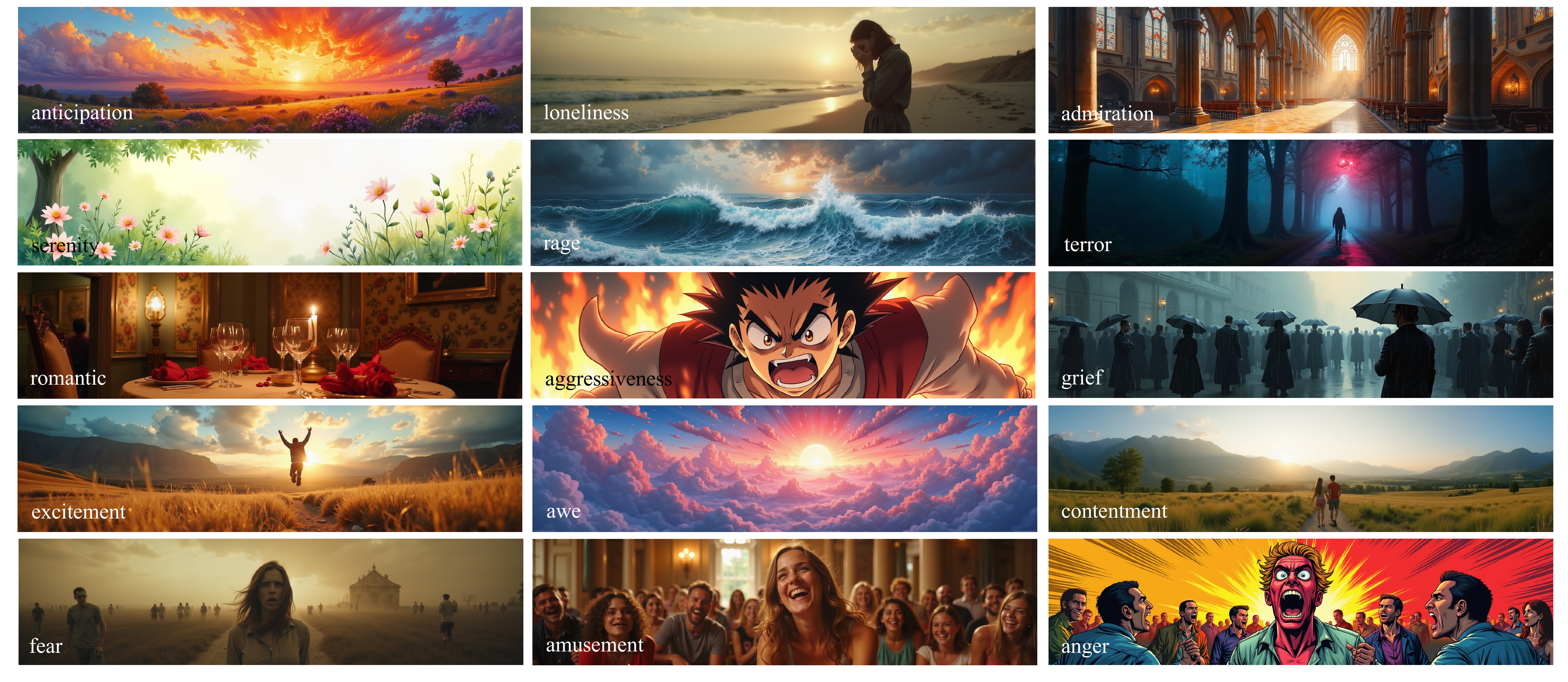}
\caption{\textbf{Example Results Generated by \textit{EmoSpace}.} We demonstrate \textit{EmoSpace}'s capability for panoramic and stylized emotional content generation combining artistic styles with fine-grained emotional control.}
\label{fig:application}
}
\end{figure*}

\begin{abstract}
\rev{
Immersive affective content generation aims to create visually compelling VR imagery with controllable emotional nuance, yet existing methods typically rely on coarse labels or prompt-only control.
Although modern diffusion transformers (DiTs) such as FLUX improve visual fidelity, they are not designed to incorporate structured affective representations. 
We present \textit{EmoSpace}, an immersive affective content generation framework guided by fine-grained emotion prototypes, transforming free-form text and emotion descriptions into fine-grained affective imagery through three coordinated components.
First, to represent sub-emotion variation beyond conventional categorical models, \textit{EmoSpace} learns a hierarchical bank of 256 prototypes with input-conditioned adaptation through vision-language alignment. 
Second, Prototype-Conditioned Steering converts these prototypes into DiT-compatible generation signals through multi-pathway injection and temporal blending, while Iterative Prompt Refinement enriches prompts with prototype-aligned sub-emotion descriptors.
Third, Affect-Grounded Modulation coordinates emotion conditioning with controllable LoRAs for panoramic, stylized, and multi-conditional generation. 
Through quantitative and qualitative evaluations, \textit{EmoSpace} improves fine-grained emotional alignment while maintaining high aesthetic quality.
Our user study shows that \textit{EmoSpace} outputs are perceived as more emotionally aligned than baseline results and more suitable for immersive scene design. Additionally, we find that immersive presentation alters emotional perception and increases emotional engagement.  
Together, these findings inform the design of emotion-aware generative systems for immersive media, with potential applications including education, immersive storytelling, and artistic creation.
We will release our code and models to facilitate future research along this line.

}
\end{abstract}

\begin{IEEEkeywords}
\rev{
Affective Computing, Diffusion Transformers, Emotion Prototype Learning, Immersive Content Generation.

}
\end{IEEEkeywords}

\section{Introduction}
\label{sec:intro}

Virtual reality (VR) has emerged as a transformative medium for creating and consuming immersive digital experiences. Emotional engagement, defined as the intensity and authenticity of users' emotional responses to visual stimuli~\cite{pavic2023feeling, somarathna2022virtual}, is essential to compelling VR experiences~\cite{luong2021survey, naz2017emotional}. It can enhance presence, immersion, and user satisfaction in diverse applications, such as entertainment, education, therapy, and training~\cite{hirsch2022traces, marin2018affective}.
With the growing demand for personalized and adaptive virtual experiences, creating emotionally rich VR content has become increasingly critical, yet remains one of the most challenging aspects in VR.

\rev{Recent advances in generative AI, particularly diffusion models, offer promising solutions for automated VR content creation~\cite{hashim2023revolutionizing, rahimi2025generative}. The field has rapidly evolved from UNet-based architectures (e.g., Stable Diffusion~\cite{podell2023sdxl}) to Diffusion Transformers (DiTs) such as FLUX~\cite{flux2024}, which improve visual fidelity through joint self-attention and flow matching. However, current methods face significant challenges in creating immersive affective content for VR environments.}

First, existing emotion modeling approaches lack the representational richness needed for fine-grained emotional expression. Such fine-grained emotion is particularly critical in VR, where immersion amplifies emotional responses~\cite{kim2014effects, kim2018effect, li2023effects}. Current approaches predominantly rely on categorical frameworks (e.g., Ekman's six basic emotions~\cite{ekman1999basic}) or dimensional models (e.g., Valence-Arousal~\cite{russell1980circumplex}). The former restrict expression to predefined labels, while the latter often lack the semantic richness required for nuanced VR experiences.

\rev{Second, current emotion-aware generation methods rely on structured inputs that limit intuitive control, creating barriers for VR content creators. Specifically, existing methods~\cite{yang2024emogen, dang2025emoticrafter} typically require predefined categories or numerical dimensions, conflicting with VR creators' workflows that involve abstract concepts, mood boards, or complex descriptions.}

\rev{Third, current emotion-aware methods primarily focus on standard 2D generation and rarely address the uniqueness of VR content, which requires generative adaptation for panoramic environments, image-conditioned generation for consistent emotional transformation, and stylized emotional expression for creative scenarios.
Additionally, the understanding of how immersive environments affect users' fine-grained emotional perception remains limited, hindering the design of emotion-aware VR content generation strategies.}

\rev{To address these challenges, we propose \textit{EmoSpace}, an immersive affective content generation framework guided by fine-grained emotion prototypes that bridges abstract emotional concepts and controllable visual synthesis within modern diffusion transformers. While prototype learning and vision-language alignment are established techniques~\cite{chen2019looks, nauta2021neural}, we adapt them to construct a hierarchical affect representation for immersive generation. Guided by psychological research~\cite{cowen2021semantic} and our formative study (see supplementary material), we observe that human emotions exhibit a ``structured-yet-continuous'' organization that is difficult to capture using either fixed categories or low-dimensional affective spaces. The central design challenge is to transform the learned hierarchy into architecture-aware conditioning signals for a modern DiT backbone and coordinate these signals with immersive-generation controls.}

\rev{Based on this insight, \textit{EmoSpace} realizes prototype-guided conditioning through three interconnected stages. First, \textbf{Emotion Prototype Learning} (Sec.~\ref{sec:proto_learning}) learns a hierarchical emotion space consisting of 256 prototypes through vision-language alignment and input-conditioned refinement, enabling fine-grained modeling of sub-emotion variation. Second, \textbf{Prototype-Guided Generation} (Sec.~\ref{sec:conditional_generation}) injects these learned prototypes into FLUX.1-dev through Prototype-Conditioned Steering, an architecture-aware mechanism that performs multi-pathway conditioning via token injection, pooled embedding modulation, and block-wise modulation with temporal blending. To complement prototype-conditioned visual guidance, Iterative Prompt Refinement enriches prompts with prototype-aligned sub-emotion descriptions. Finally, \textbf{Fine-Grained User Control} (Sec.~\ref{subsec:user_control}) extends prototype-guided conditioning to immersive applications through Affect-Grounded Modulation, coordinating prototype-conditioned modulation with external control LoRAs to support panoramic, stylized, and multi-conditional generation within a unified framework.}

\rev{To validate our approach, we conduct comprehensive evaluations covering both affective image generation (Sec.~\ref{sec:evaluation_desktop}) and immersive VR experiences (Sec.~\ref{sec:vr_user_study}). Quantitative and qualitative experiments show that \textit{EmoSpace} achieves the highest scores on the reported emotion-focused metrics, preserves aesthetic quality comparable to the strongest baseline, and receives the strongest desktop human-evaluation rankings. Within-VR comparisons further show that its affective differences from the matched FLUX+IPR baseline remain perceptible under immersive viewing and influence creative design preferences. 
Furthermore, a complementary VR-versus-desktop study indicates that, although objective recognition performance remains comparable between desktop and immersive settings, VR increases subjective emotional engagement and alters users' emotional perception patterns.}

In summary, our main contributions are:
\begin{itemize}[itemsep=0pt,topsep=0pt,parsep=0pt]
\item \rev{We propose \textit{EmoSpace}, a prototype-guided framework for immersive affective image generation that connects hierarchical fine-grained emotion representations with controllable visual synthesis.}
\item \rev{We develop Prototype-Conditioned Steering for DiT joint-attention architectures, combining prototype token injection, pooled embedding modulation, and block-wise modulation, together with Iterative Prompt Refinement for prototype-aligned prompt enhancement.}
\item \rev{We introduce Affect-Grounded Modulation for immersive multi-conditional generation and evaluate the framework through quantitative comparisons, controlled ablations, and complementary desktop and VR studies.}
\end{itemize}

\section{Related Work}
This section summarizes prior research on visual emotion representation, affective image manipulation, and immersive and affective content generation for VR.
\label{sec:related_work}

\subsection{Visual Emotion Representation}
\label{sec:emotion_model}

Visual emotion representation focuses on understanding and modeling the affective contents conveyed through visual media, ranging from psychology-grounded paradigms to data-driven spaces~\cite{wang2022systematic}.

Early approaches are grounded in two psychological paradigms: \textit{categorical models}, such as Ekman's six basic emotions~\cite{ekman1999basic} or Mikels' eight classes~\cite{mikels2005emotional}, and \textit{dimensional models}, such as the Valence-Arousal (V-A)~\cite{russell1980circumplex} and Pleasure-Arousal-Dominance (P-A-D) models~\cite{mehrabian1996pleasure}. Categorical models offer better interpretability (e.g., FI~\cite{you2015robust}), and dimensional models support finer affect intensity modeling (e.g., MAVEN~\cite{ahire2025maven}). Recent methods leverage vision-language pretraining and foundation models to enhance generalization. For example, EmoVIT~\cite{xie2024emovit} and Emotion-LLaMA~\cite{cheng2024emotion} adopt instruction-tuned vision-language transformers for emotion recognition, while GPT-4V with Emotion~\cite{lian2024gpt} and MEMO-Bench~\cite{zhou2024memo} evaluate multi-modal affective alignment in zero-shot settings.

A growing body of work explores the application of these emotional models in visual content generation. Most diffusion-based methods adopt categorical approaches and utilize emotion predictor~\cite{lin2024make}, semantic mapper~\cite{yang2024emogen}, and adapter modules~\cite{lin2024sal} to better introduce priors from emotional datasets. EmotiCrafter~\cite{dang2025emoticrafter} uses the V-A model to align different emotional categories. In visual emotion recognition, some work introduces topic guidance~\cite{chen2024tgca} or proposes counterfactual reasoning~\cite{yin2025knowledge} to improve performance and generalizability across different data distributions. \rev{However, most affective generation methods cited above are designed around earlier U-Net-based architectures (e.g., SD 1.5 and SDXL), while prototype-guided affective conditioning remains underexplored for recent DiT frameworks such as FLUX~\cite{flux2024}. Our work develops a DiT-specific prototype-conditioning mechanism and evaluates it against matched FLUX baselines. Moreover, most of these methods lack adaptive, fine-grained emotion modeling that dynamically aligns contextual semantics with emotional features. Our work bridges this gap by unifying discrete and continuous representations through learnable prototypes, supporting finer granularity, input-dependent adaptation, and a more structured prototype geometry.}

\subsection{Affective Image Manipulation}

Affective Image Manipulation (AIM) focuses on altering visual content to evoke, intensify, or transform emotional impressions~\cite{yang2025emoedit}. Prior research can be broadly categorized into three directions: emotional enhancement, emotion transfer, and emotional editing.

Emotional enhancement aims to reinforce the original affect of an image. Early works applied handcrafted adjustments to visual attributes such as hue and contrast~\cite{xu2012image}. More recent diffusion-based methods generate emotionally amplified images conditioned on textual emotion cues~\cite{zhang2025affective}. Emotion transfer involves transforming an image to reflect a different emotional state while preserving semantic content. Approaches include prompt-based fine-tuning~\cite{yang2025emoedit} and affect-adaptive latent interpolation~\cite{zhu2025bridge}. Emotional editing seeks to localize emotional changes to specific regions or stylistic elements. Some methods exploit vision-language models and valence-arousal control to guide emotional reconfiguration~\cite{mohamed2022okay, dang2025emoticrafter}.

Despite growing technical sophistication, existing approaches remain constrained by their reliance on predefined emotion categories or static dimensional frameworks~\cite{zhou2024memo}. This dependence not only entails costly annotation but also limits adaptability across diverse individual perceptions and cultural contexts. Furthermore, by treating enhancement, transfer, and editing as separate challenges, prior work has produced fragmented solutions lacking unified control mechanisms~\cite{lomas2024improved}. \rev{In contrast, our framework uses coarse category supervision to organize the training space while accepting free-form emotion descriptions at inference. The resulting structured prototype space supports multiple affective manipulation settings through a shared conditioning mechanism, without requiring a separate architecture for each task.}

\subsection{Immersive and Affective Content Generation for VR}

\rev{Recent advances in generative AI have enabled various approaches to immersive VR content creation, including static and dynamic content in 2D and 3D modalities~\cite{lee2024all, yuan2025personalized}. For panoramic imagery, diffusion models have been adapted for extended scene synthesis through specialized architectural designs, including latitude/longitude-aware mechanisms~\cite{xia2025panowan}, multi-plane synchronization~\cite{huang2025dreamcube}, and layered 3D representations~\cite{yang2025layerpano3d}. Video generation has similarly been extended to panoramic formats via spatial-temporal adaptations~\cite{liu2025dynamicscaler,zhang2024taming}. In 3D content generation, technical approaches include generating high-quality assets from 2D diffusion priors~\cite{lin2025kiss3dgen}, unbounded scene synthesis from image collections~\cite{chen2023scenedreamer}, explorable world construction from video priors~\cite{yang2025matrix}, and procedural methods for more controllable variations~\cite{raistrick2023infinite, zhao2025di}.}

While these works address technical challenges in immersive content generation, they predominantly focus on visual fidelity rather than affective design. Although VR emotion datasets~\cite{xue2021ceap,jiang2024immersive} and behavioral studies~\cite{hirsch2022traces, marin2018affective} have established methodologies for emotion measurement in virtual environments, few attempts have integrated these insights into generative pipelines. This leaves a critical gap in emotion-aware VR content generation, where emotional intent could systematically guide the generation processes, rather than treating affect as a post-hoc evaluation metric.

\section{Method}

\rev{This section presents the problem formulation and provides an overview of EmoSpace, followed by the descriptions of emotion prototype learning, prototype-guided generation, and fine-grained user control.}

\subsection{Problem Formulation}

\rev{Given the limitations of existing emotion models revealed in our formative study (see supplementary material), we formulate immersive affective content generation as follows. Let $\mathcal{I}$ denote the space of natural images and $\mathcal{E}$ a rich emotion space capturing fine-grained expressions beyond traditional categorical or dimensional constraints. Our goal is to learn a mapping $G: \mathcal{T} \times \mathcal{E} \rightarrow \mathcal{I}$ that generates emotionally coherent visual content from textual descriptions $t \in \mathcal{T}$ and emotion specifications $e \in \mathcal{E}$. The key challenges include: (1)~constructing a flexible and semantically structured emotion space $\mathcal{E}$, (2)~achieving precise emotional conditioning without compromising content quality, and (3)~extending the generation process to panoramic, stylized, and multi-conditional scenarios for VR.}

\subsection{Method Overview}
\label{subsec:method_overview}

\begin{figure*}[htbp]
\rev{
\centering
\includegraphics[width=\linewidth]{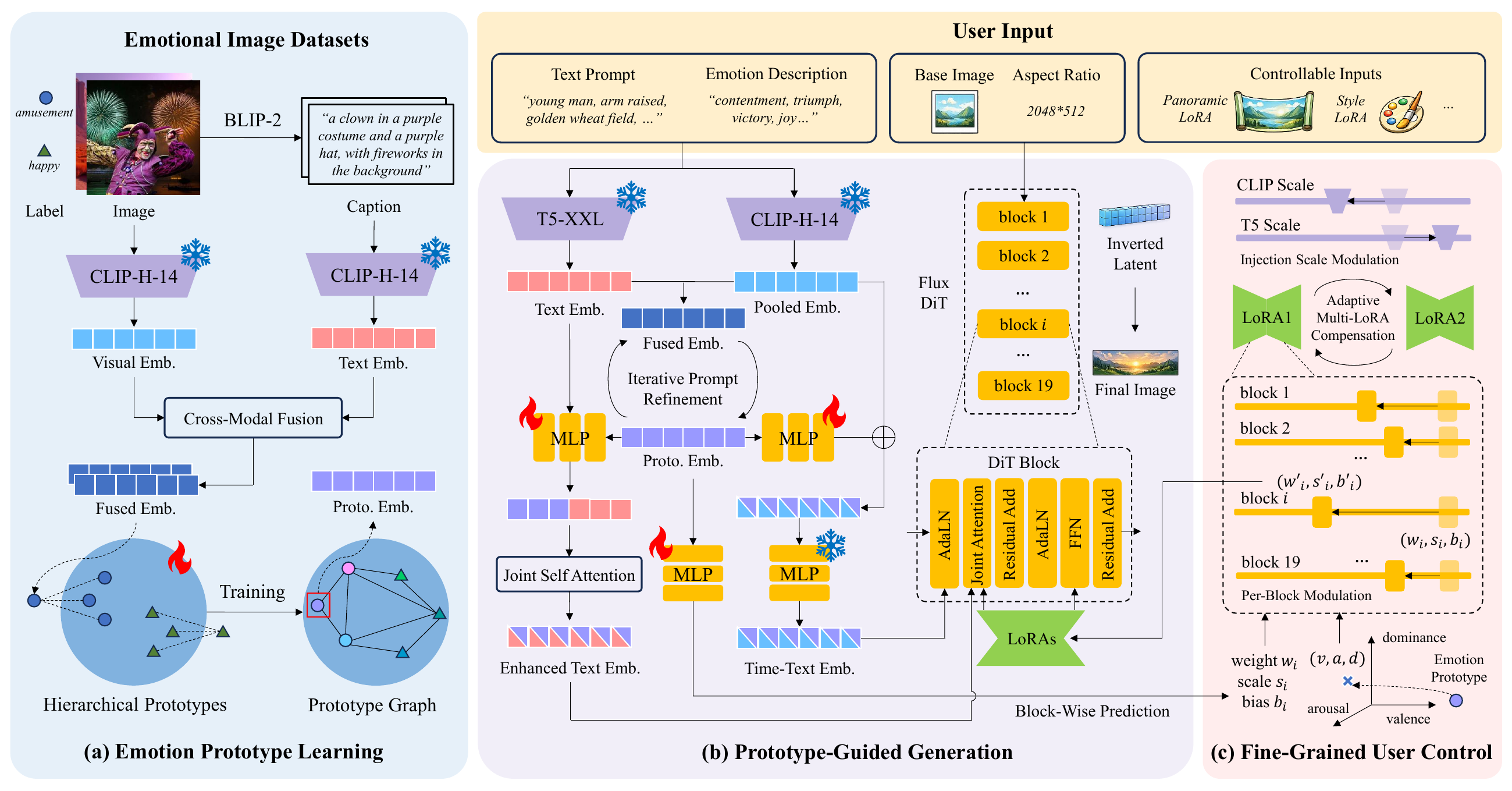}
\caption{\textbf{Framework Overview.} (a) Emotion Prototype Learning: hierarchical emotion representation with data preparation, cross-modal fusion, input-conditioned prototype adaptation, and graph-based structural refinement. (b) Prototype-Guided Generation: the learned emotion representation is converted into three FLUX conditioning pathways: prototype-token injection, pooled embedding modulation, and block-wise routing, and is complemented by sub-emotion-aware prompt refinement. (c) Fine-Grained User Control: Affect-Grounded Modulation coordinates emotion conditioning with external control LoRAs through VAD-driven pathway scaling, per-block modulation, and adaptive multi-LoRA compensation, enabling panoramic, stylized, and multi-conditional generation. Flame and snowflake icons denote trainable and frozen modules, respectively.}
\label{fig:framework}
}
\end{figure*}

As illustrated in Fig.~\ref{fig:framework}, EmoSpace addresses the above challenges through three core components:

\rev{\textbf{Emotion Prototype Learning (Sec.~\ref{sec:proto_learning}).} We construct a hierarchical prototypical emotion space that combines categorical emotion organization with fine-grained prototype discovery. The system learns 256 base prototypes and produces prototype-specific, input-conditioned adaptations, supporting structured coverage of within-category emotion variation.}

\rev{\textbf{Prototype-Guided Generation (Sec.~\ref{sec:conditional_generation}).} We develop a generation pipeline built upon FLUX.1-dev that converts the learned emotion representation into token-level, pooled, and block-wise conditioning signals through Prototype-Conditioned Steering (PCS). Iterative Prompt Refinement (IPR) complements PCS by enriching the input prompt with sub-emotion descriptors aligned with the target prototype.}

\rev{\textbf{Fine-Grained User Control (Sec.~\ref{subsec:user_control}).} Immersive applications often require emotion control to be combined with additional conditions such as style, image content, spatial structure, and panoramic layout. We introduce Affect-Grounded Modulation (AGM), which coordinates emotion conditioning with external control LoRAs through VAD-driven scale modulation and adaptive multi-LoRA compensation, supporting panoramic, stylized, and multi-conditional generation.}

\subsection{Emotion Prototype Learning}
\label{sec:proto_learning}

\rev{Prototype learning is well established for visual classification~\cite{chen2019looks, nauta2021neural}, but remains less explored for affective generation, where emotions exhibit both coarse categorical structure and substantial within-category variation. We therefore learn a hierarchical prototype space through supervised classification and contrastive representation learning.}

\textbf{Hierarchical Emotion Representation.}
\label{subsec:hierarchy}
\rev{As shown in Fig.~\ref{fig:framework}(a), the hierarchy contains $m=8$ basic emotion groups from Mikels' model~\cite{mikels2005emotional}, each with $s=32$ learnable sub-prototypes, yielding $K=256$ prototypes. Let $\mathbf{v},\mathbf{t}\in\mathbb{R}^{1024}$ denote CLIP-H-14 visual and textual embeddings~\cite{radford2021learning}. The parent emotion is predicted by}
\begin{equation}
\mathbf{y}_{\text{main}}
=
\mathbf{W}_2^{\top}
\operatorname{GELU}(\mathbf{W}_1^{\top}\mathbf{v}),
\label{eq:categorical_repr}
\end{equation}
\rev{where $\mathbf{W}_1\in\mathbb{R}^{1024\times256}$ and $\mathbf{W}_2\in\mathbb{R}^{256\times m}$. Each base prototype $\mathbf{p}_{e,j}\in\mathbb{R}^{d_p}$, with $d_p=1024$, belongs to parent group $e$ and models a local affective variation $j$. Thus, unlike flat taxonomies with a single representation per category, the hierarchy preserves interpretable parent emotions while allocating multiple learnable modes to each group. We use $K=256$ as a practical configuration rather than an optimal cardinality.}

\rev{\textbf{Dynamic Prototype Adaptation and Structural Fusion.}
During representation learning, visual and textual features are fused through
\begin{equation}
\mathbf{f}
=
\sigma\!\left(\mathbf{W}_g^f[\mathbf{v};\mathbf{t}]\right)
\odot
\mathbf{W}_f[\mathbf{v};\mathbf{t}],
\label{eq:cross_modal_fusion}
\end{equation}
where $[\cdot;\cdot]$ denotes feature concatenation.
At text-to-image inference, no target image is required. The
user-provided emotion description is encoded by the frozen
CLIP-H-14 text encoder and projected into the prototype space:
$\mathbf{f}_{\mathrm{inf}}=\psi(\mathbf{t}_{\mathrm{emo}})$,
where $\psi$ is a learned projection into the
$d_p$-dimensional prototype space. We use $\mathbf{f}_{*}=\mathbf{f}$ during representation learning and $\mathbf{f}_{*}=\mathbf{f}_{\mathrm{inf}}$ during generation.}

\rev{Rather than shifting the whole bank uniformly, two projection heads predict prototype-specific gates and offsets:
\begin{equation}
\widehat{\mathbf{p}}_{e,j}
=
\mathbf{p}_{e,j}
+
\left[
\sigma\!\left(\operatorname{reshape}(\mathbf{W}_g^o\mathbf{f}_{*})\right)
\odot
\operatorname{reshape}(\mathbf{W}_o^o\mathbf{f}_{*})
\right]_{e,j}.
\label{eq:dynamic_prototype}
\end{equation}
Let $\widehat{\mathbf{P}}(\mathbf{f}_{*})\in\mathbb{R}^{K\times d_p}$ collect the adapted prototypes. A learnable graph then refines their structural relations:
\begin{equation}
\begin{aligned}
\mathbf{M}
&=
\mathbf{A}\bigl(\widehat{\mathbf{P}}(\mathbf{f}_{*})\mathbf{W}_v\bigr),\\
\mathbf{P}'
&=
\sigma\!\left(\mathbf{A}\bigl(\widehat{\mathbf{P}}(\mathbf{f}_{*})\mathbf{W}_g\bigr)\right)
\odot\mathbf{M}
+
\widehat{\mathbf{P}}(\mathbf{f}_{*}),
\end{aligned}
\label{eq:prototype_graph}
\end{equation}
where $\mathbf{A}\in\mathbb{R}^{K\times K}$ is learned and the residual preserves the adapted bank.}

\rev{During training, the parent group is supervised by Eq.~\eqref{eq:categorical_repr}. At inference, we select the parent group by cosine similarity between $\mathbf{f}_{\mathrm{inf}}$ and each base-group centroid $\bar{\mathbf{p}}_e=s^{-1}\sum_j\mathbf{p}_{e,j}$. The corresponding graph-refined sub-bank $\mathbf{P}'_{\hat e}\in\mathbb{R}^{s\times d_p}$ is fused with the input feature through gated cross-attention:
\begin{equation}
\begin{aligned}
\mathbf{g}_a&=\sigma(\mathbf{W}_a\mathbf{f}_{*}),\\
\mathbf{f}_{\text{fused}}
&=
\mathbf{g}_a\odot
\operatorname{CrossAttn}
\!\left(
\mathbf{W}_q^a\mathbf{f}_{*},
\mathbf{P}'_{\hat e}\mathbf{W}_k^a,
\mathbf{P}'_{\hat e}\mathbf{W}_v^a
\right)
+
\mathbf{f}_{*}.
\end{aligned}
\label{eq:attention_fusion}
\end{equation}
Let $a_{\hat e,j}$ denote the normalized attention weight assigned
to sub-prototype $j$ within the predicted parent group. We define
the attention-weighted target prototype as
\begin{equation}
\mathbf{p}_{\mathrm{emo}}
=
\sum_{j=1}^{s}
a_{\hat e,j}\mathbf{p}'_{\hat e,j}.
\label{eq:weighted_target_prototype}
\end{equation}
The fused representation $\mathbf{f}_{\mathrm{fused}}$ drives the
three PCS conditioning pathways, while
$\mathbf{p}_{\mathrm{emo}}$ is used to evaluate prototype-prompt
alignment in IPR. We additionally define
$i=\arg\max_j a_{\hat e,j}$ only to index the auxiliary
sub-prototype-specific perturbations in
Eq.~\eqref{eq:diversity_amp}; the main conditioning pathway remains
a soft within-group composition.}

\rev{\textbf{Training Process.}
\label{subsec:training}
We optimize the prototype space using
\begin{equation}
\mathcal{L}
=
 w_1\mathcal{L}_{\text{main}}
+w_2\mathcal{L}_{\text{con}}
+w_3\mathcal{L}_{\text{div}}
+w_4\mathcal{L}_{\text{dist}}
+w_5\mathcal{L}_{\text{intra}}
+w_6\mathcal{L}_{\text{inter}},
\label{eq:composite_loss}
\end{equation}
with $(w_1,\ldots,w_6)=(1.0,0.5,0.1,0.1,0.5,0.3)$. The terms respectively enforce parent-emotion classification, instance-prototype alignment, prototype diversity, minimum separation, within-group diversity, and between-group separation. Complete formulations and training settings are provided in the supplementary material.}

\subsection{\rev{Prototype-Guided Generation}}
\label{sec:conditional_generation}

\rev{FLUX.1-dev~\cite{flux2024} employs 19 double-stream transformer blocks, in which text and image tokens interact through joint self-attention, followed by 38 single-stream blocks that process the merged sequence:
\begin{equation}
[\mathbf{H}_{\text{text}}',\mathbf{H}_{\text{img}}']
=
\operatorname{JointAttn}
([\mathbf{H}_{\text{text}};\mathbf{H}_{\text{img}}]),
\end{equation}
where the hidden dimension is $d=3072$. Unlike representative emotion-generation methods designed around U-Net cross-attention pathways~\cite{yang2024emogen,dang2025emoticrafter}, FLUX exposes multiple conditioning interfaces through its joint-attention architecture and pooled modulation pathway. We therefore develop an architecture-aware prototype-guided generation pipeline consisting of two complementary components, as illustrated in Fig.~\ref{fig:framework}(b).}

\rev{First, Prototype-Conditioned Steering (PCS) transforms $\mathbf{f}_{\text{fused}}$ into token-level, pooled, and block-wise conditioning signals. Prototype-conditioned routing is applied to the 19 double-stream blocks, where the text and image modalities interact explicitly, while the subsequent 38 single-stream blocks propagate the conditioned joint representation without an additional PCS routing head. Low-Rank Adaptation (LoRA)~\cite{hu2021lora} is applied to the attention and feed-forward layers for parameter-efficient emotion-specific adaptation. Second, Iterative Prompt Refinement (IPR) enriches the input prompt with sub-emotion semantics aligned with the target prototype.}

\rev{\textbf{Prototype-Conditioned Steering (PCS).}
PCS maps the fused prototype representation into three coordinated conditioning pathways and combines them with temporal blending to balance content preservation and emotional expression across the flow-matching trajectory~\cite{lipman2022flow}.}

\rev{First, we prepend $n_p=4$ learnable prototype tokens to the T5 text sequence before it enters the transformer. We compute a token gate and the corresponding emotion-token tensor:
\begin{equation}
\begin{aligned}
\mathbf{g}_{\text{tok}}
&=
\sigma(\mathbf{W}_{\text{tok}}^g\mathbf{f}_{\text{fused}}),\\
\mathbf{T}_{\text{emo}}
&=
\operatorname{reshape}\!\left(
\mathbf{g}_{\text{tok}}
\odot
f_{\text{tok}}(\mathbf{f}_{\text{fused}}),
 n_p,d
\right),\\
\widetilde{\mathbf{H}}_{\text{text}}
&=
[\mathbf{T}_{\text{emo}};\mathbf{H}_{\text{text}}].
\end{aligned}
\label{eq:token_injection}
\end{equation}
Both $f_{\text{tok}}$ and $\mathbf{W}_{\text{tok}}^g$ output $n_pd$ values, and $\mathbf{T}_{\text{emo}}\in\mathbb{R}^{n_p\times d}$. These tokens participate in joint self-attention alongside the text and image tokens. Their values are clamped to $[-1.5,1.5]$, and the final projection is zero-initialized so that training begins from the unmodified backbone behavior.}

\rev{Second, the emotion feature is projected into the pooled embedding space and added to the CLIP pooled text embedding:
\begin{equation}
\widetilde{\mathbf{c}}
=
\mathbf{c}
+
 s_p f_{\text{pool}}(\mathbf{f}_{\text{fused}}),
\label{eq:pooled_injection}
\end{equation}
where $\mathbf{c}\in\mathbb{R}^{768}$ is the pooled text embedding, $f_{\text{pool}}$ is a two-layer MLP with SiLU activation, and $s_p=0.5$. Because $\widetilde{\mathbf{c}}$ enters the AdaLayerNorm parameters through the time-text embedding pathway, it provides global emotion conditioning across the transformer.}

\rev{Third, PCS predicts block-specific routing parameters from $\mathbf{f}_{\text{fused}}$:
\begin{equation}
\begin{aligned}
\mathbf{w}_{\text{block}}
&=
\operatorname{softmax}(f_{\text{block}}(\mathbf{f}_{\text{fused}}))
&&\in\mathbb{R}^{B\times19\times n_e},\\
\mathbf{s}_{\text{block}}
&=
1.0+0.1\tanh(f_{\text{scale}}(\mathbf{f}_{\text{fused}}))
&&\in\mathbb{R}^{B\times19},\\
\mathbf{b}_{\text{block}}
&=
0.1\tanh(f_{\text{shift}}(\mathbf{f}_{\text{fused}}))
&&\in\mathbb{R}^{B\times19\times n_e}.
\end{aligned}
\label{eq:pglora}
\end{equation}
Here $n_e=2$ is the number of LoRA experts corresponding to positive and negative emotion polarity. The initial routing weights $\mathbf{w}_{\text{block}}$ describe the prototype-conditioned expert allocation, $\mathbf{s}_{\text{block}}$ provides a bounded block scale in approximately $[0.9,1.1]$, and $\mathbf{b}_{\text{block}}$ provides a bounded expert-specific adjustment to the routing logits. For block $l$ and expert $k$, the effective routing weight is
\begin{equation}
\bar{w}_{l,k}
=
\operatorname{softmax}_{k}
\left[
\log\!\left(
w_{\mathrm{block},l,k}
+
\varepsilon_{\mathrm{r}}
\right)
+
b_{\mathrm{block},l,k}
\right],
\label{eq:effective_block_routing}
\end{equation}
where $\varepsilon_{\mathrm{r}}$ is a small numerical constant. Let $\Delta\mathbf{o}_{\mathrm{emo},k}^{(l)}$ denote the residual produced by expert $k$ at double-stream block $l$. Before AGM modulation, the routed emotion residual is
\begin{equation}
\Delta\mathbf{o}_{\mathrm{emo}}^{(l)}
=
s_{\mathrm{block}}^{(l)}
\sum_{k=1}^{n_e}
\bar{w}_{l,k}
\Delta\mathbf{o}_{\mathrm{emo},k}^{(l)}.
\label{eq:routed_emotion_residual}
\end{equation}
Together, these parameters allow the emotion-modulation pattern to vary across transformer depth rather than using a single global scale.}

\rev{To align the routed emotion contribution with the generation trajectory, we use temporal blending. Following the hierarchical behavior observed in diffusion generation~\cite{dhariwal2021diffusion}, we adapt the schedule to flow matching, where the noise level $\sigma$ decreases from 1 (pure noise) to 0 (clean image). For text-conditioned generation, we employ a smooth S-curve:
\begin{equation}
 w_{\text{gen}}(\sigma)
=
\begin{cases}
0,
&\sigma>\sigma_{\text{start}},\\
\dfrac{1}{2}\!\left[
1-
\cos\!\left(
\pi
\dfrac{\sigma_{\text{start}}-\sigma}
{\sigma_{\text{start}}-\sigma_{\text{end}}}
\right)
\right],
&\sigma_{\text{end}}\leq\sigma\leq\sigma_{\text{start}},\\
1,
&\sigma<\sigma_{\text{end}},
\end{cases}
\label{eq:temporal_gen}
\end{equation}
where $\sigma_{\text{start}}=0.6$ and $\sigma_{\text{end}}=0.2$. For image-conditioned generation, we use
\begin{equation}
 w_{\text{cond}}(\sigma)
=
\exp\!\left[
-\frac{1}{2}
\frac{((1-\sigma)-\mu)^2}{\omega^2}
\right],
\end{equation}
with $\mu=0.5$ and $\omega=0.35$.}

\rev{Let $\mathbf{o}_{\text{base}}^{(l)}$ denote the unmodified output residual at double-stream block $l$, and let $\Delta\mathbf{o}_{\text{emo}}^{(l)}$ denote the additional residual contributed by the prototype-routed emotion LoRA experts. The emotion-conditioned residual and temporally blended output are
\begin{equation}
\begin{aligned}
\mathbf{o}_{\text{cond}}^{(l)}
&=
\mathbf{o}_{\text{base}}^{(l)}
+
\Delta\mathbf{o}_{\text{emo}}^{(l)},\\
\widetilde{\mathbf{o}}^{(l)}
&=
(1-w)\mathbf{o}_{\text{base}}^{(l)}
+
 w\mathbf{o}_{\text{cond}}^{(l)}
=
\mathbf{o}_{\text{base}}^{(l)}
+
 w\Delta\mathbf{o}_{\text{emo}}^{(l)},
\end{aligned}
\label{eq:temporal_blending_output}
\end{equation}
where $w$ is selected from $w_{\text{gen}}$ or $w_{\text{cond}}$ according to the generation setting.}

\rev{To preserve small differences between sub-prototypes after projection, we add auxiliary learnable perturbations indexed by the dominant within-group sub-prototype $i$ defined after Eq.~\eqref{eq:weighted_target_prototype}:
\begin{equation}
\begin{aligned}
\widehat{\mathbf{T}}_{\text{emo}}^{(i)}
&=
\mathbf{T}_{\text{emo}}
+
 s_t\boldsymbol{\delta}_t^{(i)},\\
\widehat{\Delta\mathbf{o}}_{\text{emo}}^{(l,i)}
&=
\Delta\mathbf{o}_{\text{emo}}^{(l)}
+
 s_o\boldsymbol{\delta}_o^{(l,i)}.
\end{aligned}
\label{eq:diversity_amp}
\end{equation}
Here $\boldsymbol{\delta}_t^{(i)}\in\mathbb{R}^{n_p\times d}$, while $\boldsymbol{\delta}_o^{(l,i)}$ is parameterized in the same tensor space as the routed emotion residual at block $l$. The perturbed quantities replace $\mathbf{T}_{\text{emo}}$ and $\Delta\mathbf{o}_{\text{emo}}^{(l)}$ before temporal blending. Their magnitudes are constrained to 10--20\% of the corresponding unperturbed injection norm.}

\begin{figure}[t]
\centering
\includegraphics[width=\linewidth]{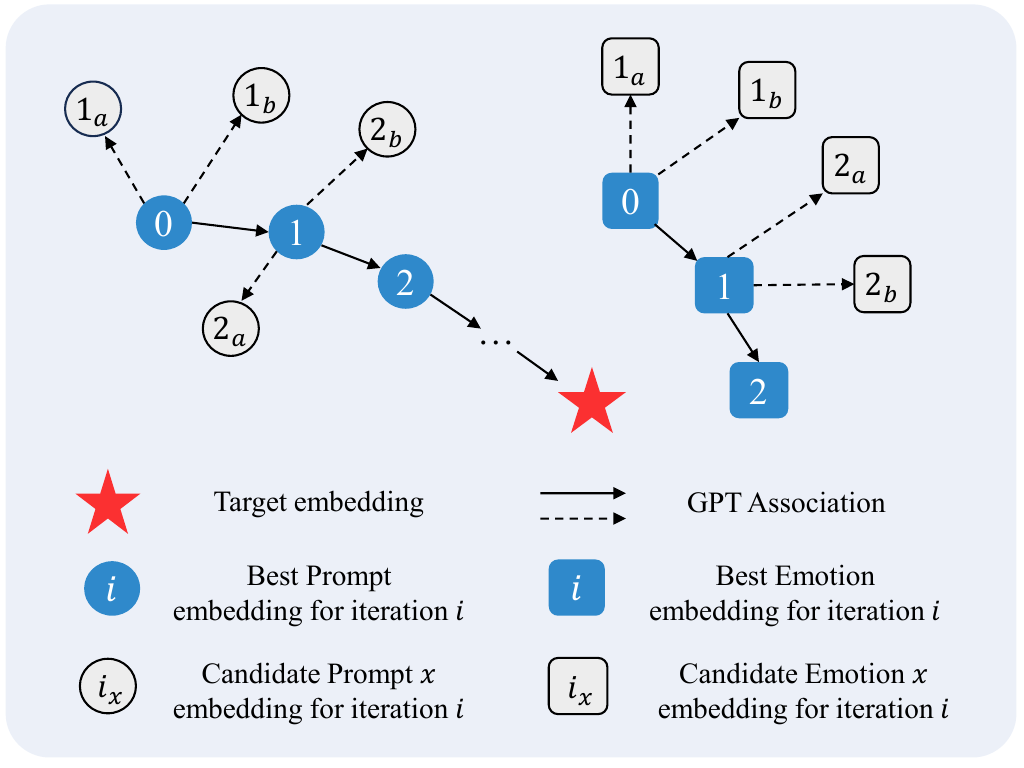}
\caption{\rev{\textbf{Iterative Prompt Refinement in the Shared Semantic Space.} Starting from the user prompt, GPT-4o-mini proposes candidate emotion descriptions and refined prompts. Their embeddings are compared with the target prototype, and the highest-scoring candidate is retained until the stopping criterion is reached.}}
\label{fig:prompt_refinement}
\end{figure}

\rev{\textbf{Iterative Prompt Refinement (IPR).}
While PCS injects prototype signals through token-level, pooled, and block-wise conditioning pathways, the input prompt may still express only coarse emotion semantics (e.g., ``anger''). IPR enriches the prompt with sub-emotion descriptors aligned with the target prototype so that the textual condition and prototype-guided generation signal convey consistent fine-grained affect. Fig.~\ref{fig:prompt_refinement} illustrates this process.}

\rev{Given an initial prompt $p_0$ and target emotion prototype $\mathbf{p}_{\text{emo}}$, GPT-4o-mini~\cite{hurst2024gpt} first proposes candidate emotion descriptions $\mathcal{C}=\{c_1,c_2,\ldots,c_n\}$ and corresponding refined prompts. We select the dominant candidate as
\begin{equation}
 c^*
=
\arg\max_{c_i\in\mathcal{C}}
\operatorname{cos}
\left(
\phi_{\text{text}}(c_i),
\mathbf{p}_{\text{emo}}
\right),
\label{eq:ipr_candidate_selection}
\end{equation}
where $\phi_{\text{text}}$ maps each candidate description into the same normalized 1024-dimensional embedding space as the emotion prototypes. The highest-scoring refined prompt is retained at each iteration. Refinement terminates when the change in the best prototype-prompt similarity satisfies
$|s^{(r)}-s^{(r-1)}|<\epsilon_{\text{ipr}}$
or when the maximum number of iterations is reached. In all reported experiments, we use $\epsilon_{\text{ipr}}=0.005$ and allow at most three refinement iterations. Detailed execution traces of two representative IPR examples are provided in the supplementary material.}

\subsection{\rev{Fine-Grained User Control}}
\label{subsec:user_control}

\rev{Immersive content creation often requires emotion control to be combined with additional conditions, such as artistic style, spatial structure, panoramic layout, and image content. Although PCS and IPR provide fine-grained prototype-conditioned emotion control, directly composing the emotion pathway with independently trained control LoRAs may weaken its relative contribution when their residuals compete.}

\rev{To support such multi-conditional settings, we introduce \textbf{Affect-Grounded Modulation (AGM)}, a lightweight coordination module built on top of PCS. AGM uses the target Valence-Arousal-Dominance (VAD) profile as an affect-grounded signal to adapt pathway- and block-level modulation and to compensate for interference from external control LoRAs. As shown in Fig.~\ref{fig:framework}(c), it consists of three core mechanisms: (i) VAD-driven injection scale modulation, (ii) VAD-decomposed per-block modulation, and (iii) adaptive multi-LoRA compensation.}

\rev{\textbf{VAD-Driven Injection Scale Modulation.}
We establish a fixed task-specific VAD lexicon that associates the parent and fine-grained emotion descriptors in our hierarchy with normalized valence, arousal, and dominance values. For a free-form emotion description, we retrieve and softly aggregate semantically related lexicon entries in the shared CLIP text space; the lexicon construction and retrieval procedure are detailed in the supplementary material.}

\rev{Let
$\mathbf{z}=[\bar{v},\bar{a},\bar{d}]^{\top}\in[-1,1]^3$
denote the resulting normalized VAD profile derived from the
free-form target emotion description and the descriptor associated
with the dominant within-group sub-prototype
$i$~\cite{mehrabian1996pleasure}. A lightweight modulation head converts this profile into bounded pathway scales:
\begin{equation}
\boldsymbol{\gamma}
=
\mathbf{1}
+
\rho\tanh(\mathbf{W}_{\gamma}\mathbf{z}+\mathbf{b}_{\gamma})
=
[\gamma_{\mathrm{tok}},\gamma_{\mathrm{pool}},\gamma_{\mathrm{blk}}]^{\top},
\label{eq:vad_path_scales}
\end{equation}
where $\rho$ bounds the maximum deviation from the base scale. We apply $\gamma_{\mathrm{tok}}$ to $\mathbf{T}_{\mathrm{emo}}$, $\gamma_{\mathrm{pool}}$ to the pooled offset in Eq.~\eqref{eq:pooled_injection}, and $\gamma_{\mathrm{blk}}$ to the routed block residual.}

\rev{\textbf{VAD-Decomposed Per-Block Modulation.}
AGM further modulates the PCS block scale using the individual VAD dimensions:
\begin{equation}
\widetilde{\mathbf{s}}_{\text{block}}^{(l)}
=
\mathbf{s}_{\text{block}}^{(l)}
\left[
1.0
+
\sum_{k\in\{V,A,D\}}
\alpha_{l,k}z_k
\tanh\!\left(f_k^{(l)}(\mathbf{f}_{\text{fused}})\right)
\right],
\label{eq:plm_vad}
\end{equation}
where $z_k$ is the corresponding component of $\mathbf{z}$, $f_k^{(l)}$ predicts a prototype-dependent modulation term, and $\alpha_{l,k}$ is a learned block- and dimension-specific coefficient. Thus, the target affect profile and prototype feature jointly adapt the PCS scale at each depth. When AGM is active, Eq.~\eqref{eq:routed_emotion_residual} is evaluated using $\gamma_{\mathrm{blk}}\widetilde{\mathbf{s}}_{\text{block}}^{(l)}$ in place of $s_{\text{block}}^{(l)}$; we retain $\Delta\mathbf{o}_{\text{emo}}^{(l)}$ to denote this VAD-modulated but uncompensated emotion residual.}

\rev{\textbf{Adaptive Multi-LoRA Compensation.}
When emotion conditioning is combined with $Q$ external control LoRAs, strong control residuals may attenuate the relative contribution of the emotion pathway. Let
$\widehat{\Delta\mathbf{o}}_{\mathrm{emo}}^{(l,i)}$
denote the VAD-modulated but uncompensated emotion residual at
block $l$, and let $\Delta\mathbf{o}_q^{(l)}$ denote the residual
contributed by external LoRA $q$. We define
\begin{equation}
r_{\mathrm{emo}}^{(l)}
=
\left\|
\widehat{\Delta\mathbf{o}}_{\mathrm{emo}}^{(l,i)}
\right\|_2,
\qquad
e^{(l)}
=
\left\|
\sum_{q=1}^{Q}
\Delta\mathbf{o}_q^{(l)}
\right\|_2.
\end{equation}
We then compute the bounded compensation gain
\begin{equation}
 g^{(l)}
=
1.0
+
\beta
\frac{e^{(l)}}
{r_{\mathrm{emo}}^{(l)}+e^{(l)}+\varepsilon_{\mathrm{c}}},
\label{eq:plm_compensation}
\end{equation}
where $\beta$ controls the maximum compensation strength and $\varepsilon_{\mathrm{c}}$ ensures numerical stability. The compensated emotion residual is
\begin{equation}
\widetilde{\Delta\mathbf{o}}_{\mathrm{emo}}^{(l,i)}
=
g^{(l)}
\widehat{\Delta\mathbf{o}}_{\mathrm{emo}}^{(l,i)}.
\label{eq:compensated_emo_output}
\end{equation}
The gain remains close to 1 when external controls are weak and approaches $1+\beta$ as their aggregate residual becomes dominant. Together with the VAD-conditioned pathway and block scales, this bounded rescaling supports balanced composition of emotion and external control LoRAs.}

\section{Evaluation of Content Generation}

\label{sec:evaluation_desktop}

\rev{This section presents our evaluation of emotional content generation quality, including comparative experiments, ablation studies, and human evaluation.}

\subsection{\rev{Application Scenarios}}
\label{subsec:application_scenarios}


\rev{Building on AGM for multi-LoRA coordination, we extend the \textit{EmoSpace} framework to address three unique requirements for immersive affective content generation: emotional panorama generation for immersive environments, stylized emotional generation with emotion-style collaborative control, and multi-conditional emotional content generation (e.g., text, image, style, resolution) for extended controllability.}

\begin{figure}[t]
\rev{
    \centering
    \includegraphics[width=\linewidth]{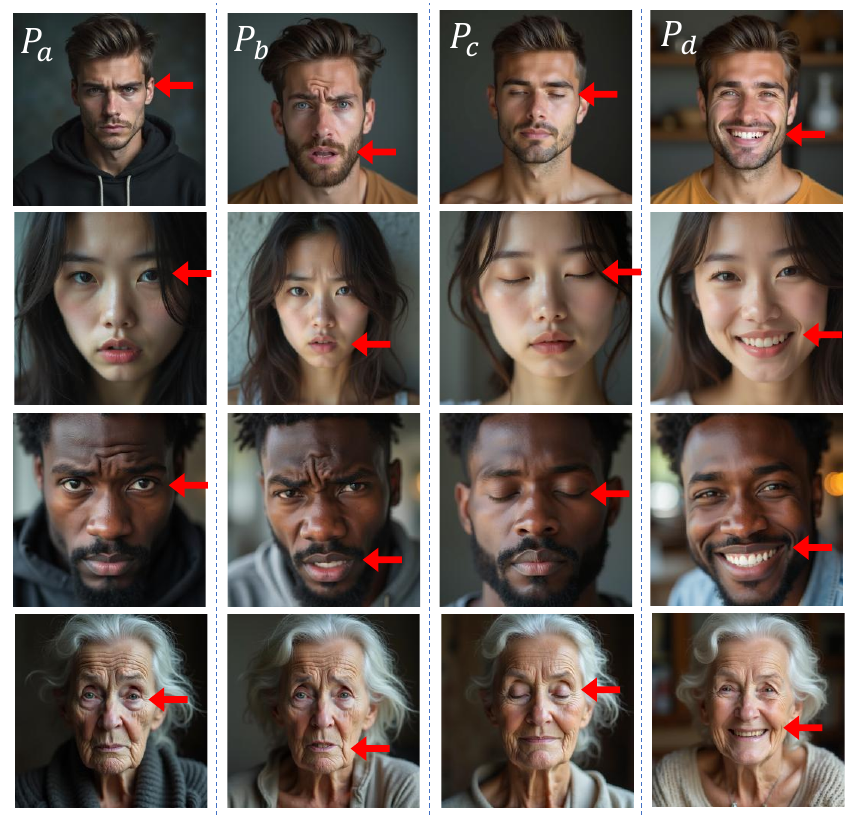}
    \caption{\textbf{Fine-Grained Emotion Separation from Sub-Prototypes.} Four sub-prototypes (Pa--Pd) generate images from different identity prompts. Pa and Pb are drawn from the \textit{anger} category, while Pc and Pd are drawn from the \textit{contentment} category. Despite sharing the same parent emotion, the sub-prototypes yield different expressions and atmospheric cues, illustrating within-category variation in the learned conditioning space.}
    \label{fig:prototype}
}
\end{figure}

\begin{figure}[t]
\rev{
    \centering
    \includegraphics[width=\linewidth]{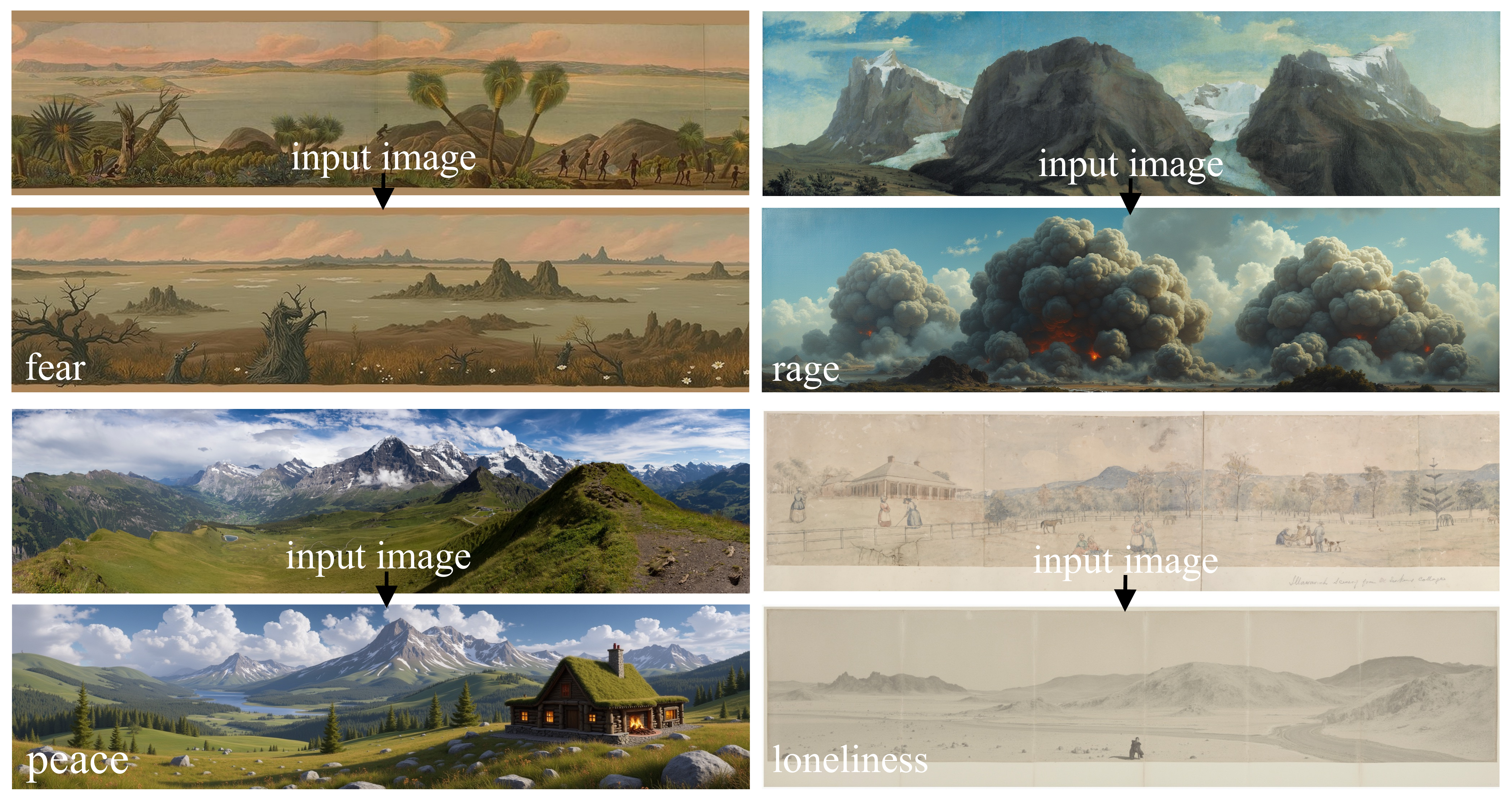}
    \caption{\textbf{Image-Conditioned Panorama Generation with Emotion Control.} The reference image provides content and structural cues, while prototype-guided conditioning controls the target affect during panoramic generation.}
    \label{fig:image_based}
}
\end{figure}

\rev{Fig.~\ref{fig:application} provides an overview of the generated panoramas and stylized scenes. Across the displayed styles and emotional descriptions, the affective cues remain visible while scene content and visual style vary, illustrating the intended separation between emotion control and other generation conditions. More examples are provided in the supplementary material. Fig.~\ref{fig:prototype} further shows within-category variation: sub-prototypes under the same parent emotion produce distinct expressions and atmospheric cues across different identity prompts. Fig.~\ref{fig:image_based} demonstrates image-conditioned panoramic generation, where the reference layout is preserved while emotion and panoramic format are jointly controlled. Table~\ref{tab:comparison} summarizes the corresponding functional scope: general SDXL and FLUX models support the underlying generation tasks, whereas \textit{EmoSpace} additionally provides explicit categorical and fine-grained affect control on a DiT backbone. Together, these examples show how the proposed conditioning pathway supports semantic, stylistic, image-conditioned, and panoramic use cases within one framework.}

\begin{table}[t]
\rev{
\centering
\small
\setlength{\tabcolsep}{4pt}
\caption{\textbf{Functional Comparison.} Comparison of supported generative tasks and emotion control approaches across different methods. T2I: text-to-image generation, ImgCond: image-conditioned generation, Cat.: emotional control via categorical labels, FG: fine-grained emotion control, DiT: built on a diffusion-transformer architecture. \checkmark: supported, \ding{55}: not supported.}
\label{tab:comparison}
\resizebox{\columnwidth}{!}{%
\begin{tabular}{lccccccc}
\toprule
\textbf{Method} & \textbf{T2I} & \textbf{ImgCond} & \textbf{Style} & \textbf{Panorama} & \textbf{Cat.} & \textbf{FG} & \textbf{DiT} \\
\midrule
EmoGen & \checkmark & \ding{55} & \ding{55} & \ding{55} & \checkmark & \ding{55} & \ding{55} \\
EmotiCrafter & \checkmark & \ding{55} & \ding{55} & \ding{55} & \ding{55} & \checkmark & \ding{55} \\
SDXL & \checkmark & \checkmark & \checkmark & \checkmark & \ding{55} & \ding{55} & \ding{55} \\
FLUX & \checkmark & \checkmark & \checkmark & \checkmark & \ding{55} & \ding{55} & \checkmark \\
\textbf{EmoSpace (Ours)} & \checkmark & \checkmark & \checkmark & \checkmark & \checkmark & \checkmark & \checkmark \\
\bottomrule
\end{tabular}}
}
\end{table}

\subsection{Experimental Settings}

\rev{\textbf{Datasets.} We use images and emotion labels from EmoSet-118K~\cite{yang2023emoset} for training, chosen for its large scale (118K images) and alignment with Mikels' 8-emotion taxonomy that matches our prototype hierarchy. Since this dataset lacks textual descriptions, we employ BLIP-2~\cite{li2023blip} to generate captions for all images, enabling the training process in Sec.~\ref{sec:proto_learning}. For evaluation, we generate 25 images per emotion category (200 in total) using each method and compute metrics on the generated images.}

\rev{\textbf{Baselines.} We compare against prompt-based and conditioning-based methods. The prompt-based baselines are SDXL~\cite{podell2023sdxl}, FLUX.1-dev~\cite{flux2024}, and FLUX.2-dev~\cite{flux2}, a newer DiT model with a larger architecture and a Mistral-based text encoder. SDXL, FLUX.1-dev, and FLUX.2-dev receive the same IPR-refined prompts as EmoSpace, allowing the comparison with FLUX+IPR to isolate prototype-guided conditioning from prompt refinement. The conditioning-based methods are EmoGen~\cite{yang2024emogen}, which uses learnable emotion tokens and LLM-based prompt optimization, and EmotiCrafter~\cite{dang2025emoticrafter}, which injects VAD embeddings with continuous control; both are SDXL-based. For these methods, we provide the richest conditioning supported by their original interfaces: class labels for EmoGen and estimated valence-arousal values for EmotiCrafter. }

\rev{\textbf{Evaluation Metrics.} To evaluate the generated results, we use complementary measures covering emotion alignment, separability, perceptual differentiation, visual quality, and human judgment. Emotion Accuracy (EA), Fine-grained Emotion Accuracy (FgEA), Emotion Separability Score (ESS), and Sub-Emotion Discriminability (SED) are computed with a held-out OpenCLIP evaluator that shares no learned parameters with EmoSpace. These metrics reduce direct optimization leakage, but they remain vision-language-space measures and therefore do not fully eliminate representation coupling. Cross-LPIPS~\cite{zhang2018unreasonable} provides a non-CLIP measure of perceptual differentiation across emotions, while LAION-Aesthetics~\cite{schuhmann2022laion} assesses visual quality rather than affective correctness. We therefore interpret automatic results together with controlled ablations and human judgments of basic and fine-grained emotional accuracy. Detailed definitions and limitations are provided in the supplementary material.}

\rev{\textbf{Implementation Details.} Our framework is built upon FLUX.1-dev with 12B parameters. Stage 1 trains the emotion encoder with CLIP-H-14 backbone (frozen) for 20 epochs with batch size 16 and learning rate $5 \times 10^{-5}$. Stage 2 trains the prototype injector, block-wise modulation predictor, and LoRA parameters (rank=32, $\alpha$=8) for 10,000 steps with batch size 1 and gradient accumulation of 4. The diversity amplification module uses token perturbation strength 3.0 and routed-residual perturbation strength 0.5. All experiments use bfloat16 precision on a single NVIDIA RTX 6000 Ada GPU (48GB VRAM). The additional trainable parameters (injector, modulation predictor, amplifier, prototype embeddings, and LoRA adapters) total 122.70M, corresponding to 1.02\% of the FLUX transformer. Under a controlled profile at $512\times512$ resolution and 30 denoising steps, Base FLUX requires $10.80\pm0.20$~s per image, while the full EmoSpace generation model, excluding the external IPR call, requires $10.94\pm0.20$~s, corresponding to an overhead of approximately $0.14$~s (1.3\%); peak VRAM is 30.49~GB for both configurations. Detailed profiling is provided in the supplementary material.}

\subsection{\rev{Comparative Study}}

\rev{\textbf{Quantitative Results.} Table~\ref{tab:quantitative} presents the quantitative comparison. EmoSpace obtains the highest value on all five emotion-focused metrics: \textbf{ESS} (\textbf{0.616} vs. 0.554 for FLUX.2-dev), \textbf{EA} (\textbf{0.633} vs. 0.610), \textbf{FgEA} (\textbf{0.645} vs. 0.615), \textbf{SED} (\textbf{0.901} vs. 0.843), and \textbf{Cross-LPIPS} (\textbf{0.721} vs. 0.696). ESS uses macro-F1 to penalize category imbalance; EmoSpace's higher ESS therefore reflects stronger class-balanced separation under the reported evaluator. For \textbf{LAION-Aesthetics}, EmoSpace (5.970) is comparable to the strongest baseline, EmotiCrafter (5.984), indicating that the added affective control preserves the reported aesthetic quality. The matched FLUX+IPR comparison is particularly informative: with identical refined prompts, adding prototype-guided conditioning increases ESS from 0.514 to 0.616, EA from 0.575 to 0.633, FgEA from 0.615 to 0.645, SED from 0.843 to 0.901, and Cross-LPIPS from 0.655 to 0.721. These results support the complementary contribution of the visual prototype pathway beyond prompt refinement, while the aesthetic result is interpreted as preservation rather than superiority in visual quality.}

\begin{table}[t]
\rev{
\centering
\small
\setlength{\tabcolsep}{3pt}
\caption{\textbf{Quantitative Comparison.} SDXL, FLUX.1-dev, and FLUX.2-dev receive the same IPR-refined prompts as EmoSpace; EmoGen and EmotiCrafter receive the richest emotion conditions supported by their original interfaces. EmoSpace differs from the matched FLUX+IPR baseline through prototype-guided conditioning. The table reports five emotion-focused metrics (ESS, EA, FgEA, SED, Cross-LPIPS) plus one aesthetic metric (LAION-Aesth). ESS: Emotion Separability Score (macro-F1 of zero-shot CLIP classification, class-balanced), EA: zero-shot Emotion Accuracy, FgEA: Fine-grained EA (top-5 sub-emotion majority voting), SED: Sub-Emotion Discriminability, Cross-LPIPS: cross-emotion perceptual diversity, LAION-Aesth: score from the officially released LAION-Aesthetics predictor. Best results in \textbf{bold}. $\dagger$: second best.}
\label{tab:quantitative}
\resizebox{\columnwidth}{!}{%
\begin{tabular}{lcccccc}
\toprule
\textbf{Method} & \textbf{ESS}$\uparrow$ & \textbf{EA}$\uparrow$ & \textbf{FgEA}$\uparrow$ & \textbf{SED}$\uparrow$ & \textbf{Cross-LPIPS}$\uparrow$ & \textbf{LAION-Aesth}$\uparrow$ \\
\midrule
SDXL+IPR & 0.263 & 0.290 & 0.295 & 0.838 & 0.674 & 5.365 \\
EmoGen & 0.068 & 0.115 & 0.175 & 0.840 & 0.696$^\dagger$ & 5.087 \\
EmotiCrafter & 0.303 & 0.315 & 0.430 & 0.826 & 0.576 & \textbf{5.984} \\
FLUX+IPR & 0.514 & 0.575 & 0.615$^\dagger$ & 0.843$^\dagger$ & 0.655 & 5.906 \\
FLUX.2+IPR & 0.554$^\dagger$ & 0.610$^\dagger$ & 0.610 & 0.827 & 0.627 & 5.702 \\
\textbf{EmoSpace (Ours)} & \textbf{0.616} & \textbf{0.633} & \textbf{0.645} & \textbf{0.901} & \textbf{0.721} & 5.970$^\dagger$ \\
\bottomrule
\end{tabular}}
}
\end{table}
\textbf{Category-Level Analysis.} The supplementary material reports mean $\pm$ standard deviation across the same eight emotion categories for all six methods, including FLUX.2+IPR. Because these categories are fixed, matched evaluation conditions rather than independent random samples, we do not treat the eight category summaries as independent replicates for an omnibus significance test. Instead, we report category-level dispersion and matched effect magnitudes. Relative to FLUX+IPR, EmoSpace improves ESS by 0.102, EA by 0.058, FgEA by 0.030, and SED by 0.058; relative to FLUX.2+IPR, the corresponding gains are 0.062, 0.023, 0.035, and 0.074. EmoSpace also has the lowest category-level standard deviation for ESS and SED, indicating more consistent performance across the evaluated emotions. Cross-LPIPS and LAION-Aesthetics are aggregate metrics and are reported descriptively.

\begin{figure*}[htbp]
\rev{
\centering
\includegraphics[width=\linewidth]{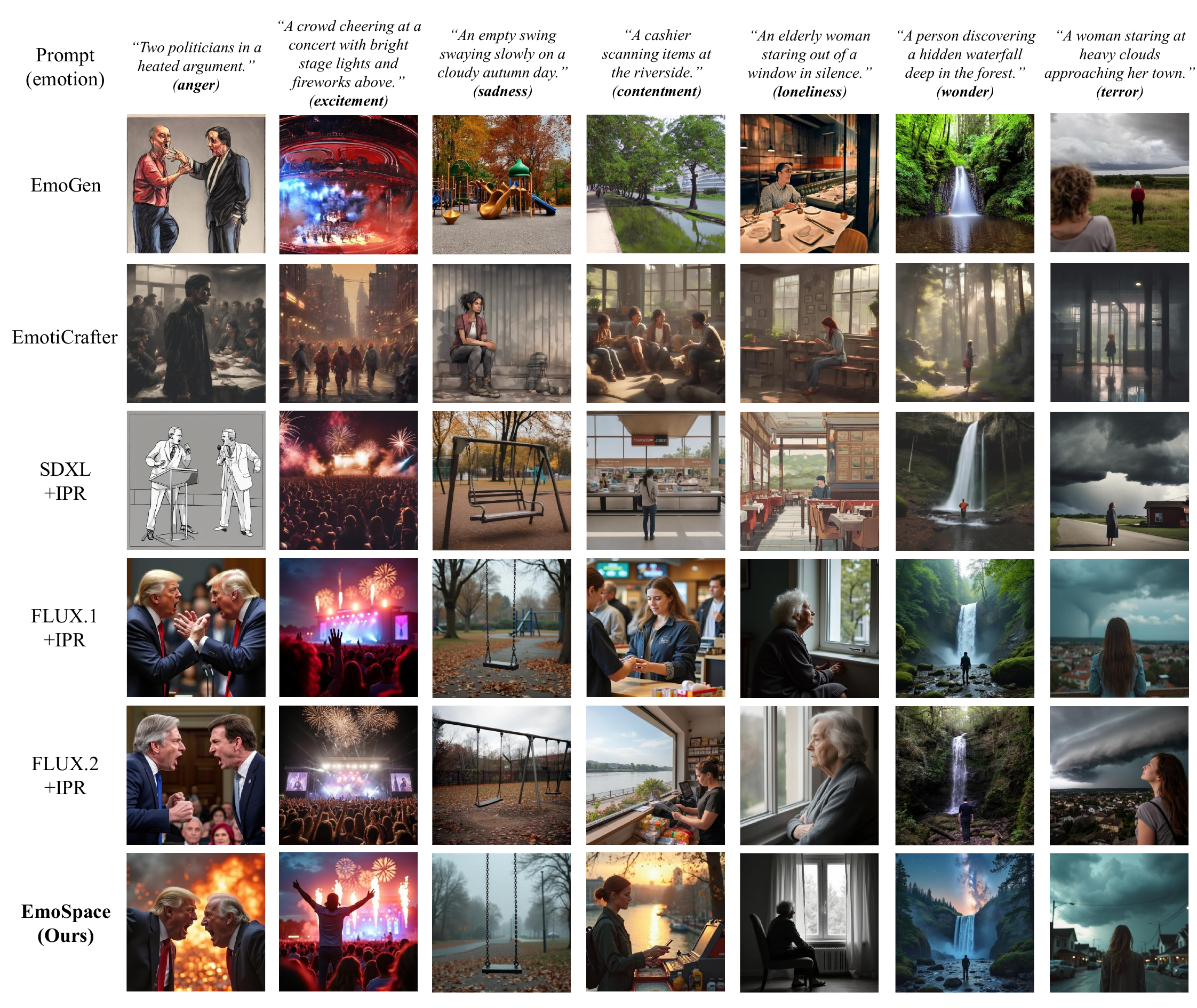}
\caption{\textbf{Qualitative Comparison.} Generated images across seven emotions for six methods: EmoGen, EmotiCrafter, SDXL+IPR, FLUX.1+IPR, FLUX.2+IPR, and EmoSpace (Ours). In these examples, \textit{EmoSpace} produces more distinct and recognizable emotional expressions, particularly for subtle emotions such as \textit{loneliness} and \textit{wonder}.}
\label{fig:comparison}
}
\end{figure*}

\rev{\textbf{Qualitative Results.} Fig.~\ref{fig:comparison} shows qualitative comparisons across the evaluated methods and 7 different emotions. According to the examples, \textit{EmoSpace} exhibits more differentiated emotional expressions while retaining comparable or better visual fidelity. Several baselines show weaker distinctions in these examples for subtle emotions such as \textit{contentment}, \textit{loneliness}, and \textit{wonder}. These qualitative observations further support the conclusions from the quantitative evaluations.}

\subsection{Ablation Study}
\label{subsec:ablation}

\rev{We conduct comprehensive ablation studies to validate each component's contribution. All ablations use the same FLUX backbone and training protocol.}

\rev{\textbf{Quantitative Results.} 
Table~\ref{tab:ablation} reports system-level ablations aligned with PCS, IPR, and AGM. We use the same five metrics as in Table~\ref{tab:quantitative} for direct comparison. Each removal is associated with lower performance on multiple metrics. Removing IPR produces the largest reduction in FgEA and Cross-LPIPS, indicating that prototype-aligned prompt enrichment contributes strongly to fine-grained textual guidance. Removing PCS lowers ESS, EA, and FgEA, supporting the contribution of DiT-specific prototype steering beyond the matched prompt pathway. Removing AGM lowers FgEA and SED, suggesting that adaptive modulation is useful when emotion conditioning is composed with additional controls. The gradual degradation is consistent with the complementary design of the three pathways, whose effects accumulate across textual, prototype, and multi-control conditioning.}

\begin{table}[t]
\rev{
\centering
\small
\setlength{\tabcolsep}{3pt}
\caption{\textbf{Component Ablation.} Removing each component from the full model, aligned with our three core contributions (PCS, IPR, AGM). All metrics match the comparison table (Table~\ref{tab:quantitative}). The table reports descriptive system-level changes under the same evaluation protocol.}
\label{tab:ablation}
\resizebox{\columnwidth}{!}{%
\begin{tabular}{lccccc}
\toprule
\textbf{Configuration} & \textbf{ESS}$\uparrow$ & \textbf{EA}$\uparrow$ & \textbf{FgEA}$\uparrow$ & \textbf{SED}$\uparrow$ & \textbf{Cross-LPIPS}$\uparrow$ \\
\midrule
Full Model & \textbf{0.616} & \textbf{0.633} & \textbf{0.645} & \textbf{0.901} & \textbf{0.721} \\
w/o IPR & 0.559 & 0.625 & 0.465 & 0.736 & 0.601 \\
w/o PCS & 0.595 & 0.620 & 0.630 & 0.766 & 0.702 \\
w/o AGM & 0.571 & 0.600 & 0.630 & 0.759 & 0.699 \\
\bottomrule
\end{tabular}}
}
\end{table}

\begin{figure}[htbp]
\rev{
\centering
\includegraphics[width=\columnwidth]{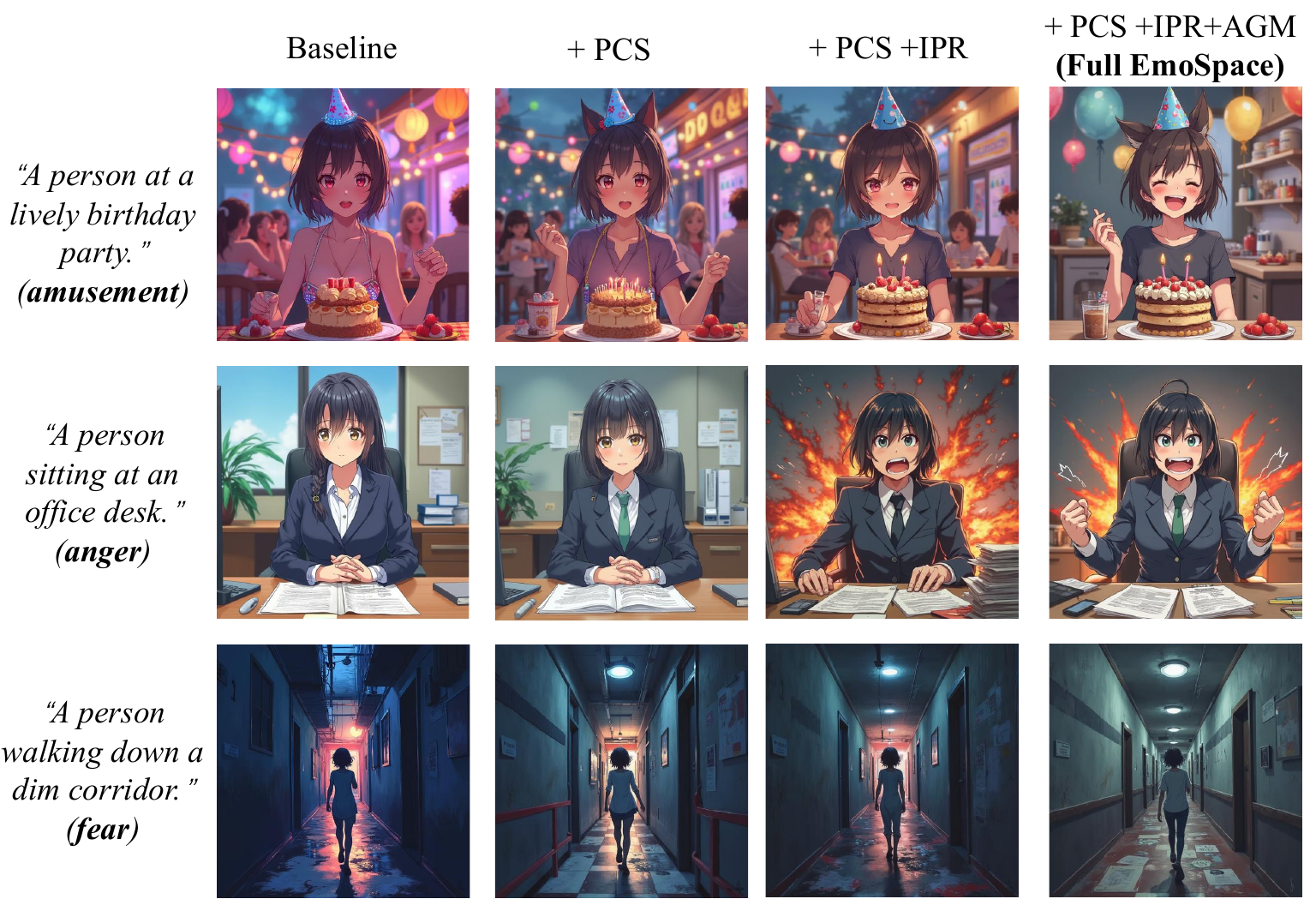}
\caption{\textbf{Ablation Study.} Qualitative comparison of the full model and three ablated configurations for representative emotions. The examples visualize changes in facial expression, gesture, lighting, and environmental atmosphere when IPR, PCS, or AGM is removed.}
\label{fig:ablation}
}
\end{figure}

\rev{\textbf{Qualitative Results.} Fig.~\ref{fig:ablation} provides a qualitative demonstration of component ablation. According to the results, PCS strengthens prototype-driven differences in expression and atmosphere; IPR introduces specific emotional cues in the prompt-dependent details; and AGM increases the consistency of emotional cues when multiple controls are combined. The full configuration shows the clearest joint expression of facial, gestural, and environmental signals in the displayed examples.}

\rev{\textbf{Prototype Analysis.} We analyze the learned bank at both representation and generation levels. In the original CLIP space, 210 of 256 base prototypes are covered by held-out validation features, with normalized entropy 0.752, indicating broad bank-level coverage. A controlled dynamic-versus-static comparison further yields higher values with input-conditioned adaptation (EA 0.633 vs.\ 0.598, FgEA 0.645 vs.\ 0.620, SED 0.901 vs.\ 0.870, and Cross-LPIPS 0.721 vs.\ 0.685). Together, these results characterize the learned bank as a structured conditioning space with broad raw-space coverage and measurable benefits from input-conditioned adaptation. Full definitions and diagnostics are provided in the supplementary material.}

\subsection{Human Evaluation}
\label{subsec:human_eval}

\rev{We conducted a desktop ranking study comparing EmoSpace with five baselines (SDXL+IPR, FLUX.2+IPR, FLUX+IPR, EmoGen, and EmotiCrafter) across 26 matched emotional prompts. Each prompt set contained one image from each of the six methods, yielding 156 generated images in total. We recruited 28 participants (14 male, 14 female, ages 20--35) for this evaluation.}

\rev{For each prompt set, participants viewed the six randomly ordered images side by side and ranked them from 1 (best) to 6 (worst) on four dimensions: (1) text-image alignment, (2) basic emotion accuracy, (3) fine-grained emotion accuracy, and (4) aesthetic quality. Rankings were averaged within participant across the 26 prompt sets before aggregation, so repeated prompt-level judgments were not treated as independent observations. We therefore report participant-aggregated mean ranks descriptively.}

\rev{As shown in Fig.~\ref{fig:task_performance}(a), EmoSpace obtains the lowest (best) mean rank across all four dimensions: \textbf{text-image alignment} (1.83), \textbf{basic emotion accuracy} (1.88), \textbf{fine-grained emotion accuracy} (1.85), and \textbf{aesthetic quality} (2.05). FLUX.2+IPR is the second-ranked method across the four criteria (mean ranks 2.42--2.55). The consistent ordering complements the automatic evaluation, while the qualitative examples suggest that the distinction is particularly visible for subtle emotions such as \textit{awe} and \textit{contentment}.}

\section{Evaluation of VR Experience}

\label{sec:vr_user_study}

\rev{Our focus on immersive affective experiences necessitates evaluation beyond traditional desktop settings. We design three experiments to comprehensively evaluate EmoSpace in VR contexts: (1) a \textit{method-specific} within-VR comparison, (2) a \textit{creative design preference} study, and (3) a \textit{medium comparison}.}

\subsection{User Study Protocol}
\label{subsec:vr_setting}

\rev{\textbf{Stimuli.} Stimuli for Experiments 1 and 2 are generated using the FLUX.1-dev backbone, while Experiment 3 uses stimuli from our SDXL-based implementation (details and visual effects in supplementary materials). We use four stimulus sets of panoramic images (2048$\times$512, rendered at 240°$\times$60° FoV in VR). The first set consists of 32 EmoSpace panoramas (8 basic emotions $\times$ 4 Plutchik sub-emotions) generated with prototype-guided injection and iterative prompt refinement. The second set contains 32 FLUX+IPR panoramas generated from the same IPR-refined prompts but without prototype injection, serving as the architecture-controlled baseline. From these two matched stimulus pools, we selected 20 panorama pairs for Experiment~1, balancing the represented basic and fine-grained emotion conditions. The third set contains eight task-oriented panoramas (four matched method pairs) used in the creative design study. The fourth set contains 16 SDXL-generated panoramas, each presented identically in VR and desktop for the medium comparison. Experiments 1 and 2 isolate the \textit{method} effect by comparing EmoSpace with a matched FLUX+IPR baseline within the same VR display condition. Experiment 3 isolates the \textit{medium} effect by presenting identical SDXL-generated stimuli in VR and desktop conditions. The use of identical stimuli, rather than the choice of backbone itself, supports the medium comparison. The SDXL adaptation additionally provides qualitative evidence that the conditioning paradigm can be implemented on both DiT and UNet backbones, with backbone-specific injection mechanisms described in the supplementary material.}

\rev{\textbf{Procedure.} 
The study consists of three experiments conducted in a single session (approximately 45 minutes):}

\rev{\textit{Experiment 1: Method-Specific Comparison (within VR).} 
Each participant views 20 paired panoramas (EmoSpace vs.\ FLUX+IPR) from identical emotional prompts in randomized order. For each pair, participants complete: (\textbf{T1}) \textit{Emotion Accuracy}, which image better expresses the target emotion (forced choice); (\textbf{T2}) \textit{Emotional Intensity}, rating each image separately on a 7-point Likert scale; and (\textbf{T3}) \textit{Fine-grained Discrimination}, which image offers richer emotional nuance and subtlety (forced choice). This experiment tests whether the full \textit{EmoSpace} framework yields perceptible affective differences within the same VR environment, thereby holding the presentation medium constant.}

\rev{\textit{Experiment 2: Creative Design Preference Study (within VR).} 
To evaluate practical utility in creative workflows, participants are given a creative scenario: ``You are designing an immersive VR environment for a specific emotional experience.'' For each of 4 target emotions, participants: (a) view the EmoSpace-generated panorama and the FLUX+IPR panorama side-by-side in VR, (b) select which environment they would choose for their VR experience, and (c) rate \emph{each} environment separately on three 7-point scales: \textit{emotional alignment} (how well it matches their emotional intent), \textit{immersive potential} (how effectively it would engage a VR user), and \textit{creative suitability} (how useful it would be as a starting point for further design). This separation between forced choice and per-candidate ratings enables paired method comparisons while retaining the design-oriented selection task.}

\rev{\textit{Experiment 3: Medium Comparison (VR vs.\ Desktop).}
The same SDXL-generated stimuli are presented in both VR and desktop conditions, so the within-participant comparison isolates the presentation-medium effect. Method-specific differences are evaluated separately using the matched FLUX configurations in Experiments~1 and~2. Each participant views the same 16 panoramas once in VR and once on desktop (32 presentations in total), with condition order counterbalanced and within-condition order randomized. For each panorama, participants complete: (\textbf{T1}) \textit{Emotion Category Recognition} from 8 categories, (\textbf{T2}) \textit{Emotional Activation Rating} on a 3-point scale, and (\textbf{T3}) \textit{Fine-grained Emotion Discrimination} from 32 specific emotion words. After completing both presentation conditions, participants rate their perceived accuracy for T1, T2, and T3 separately for VR and desktop on 7-point scales (1 = very inaccurate, 7 = very accurate). A post-experience questionnaire then evaluates VR vs.\ desktop across five dimensions including immersive quality, emotional engagement, perceptual quality, cognitive load, and overall preference.}

\begin{figure}[t]
\rev{
\centering
\includegraphics[width=\linewidth]{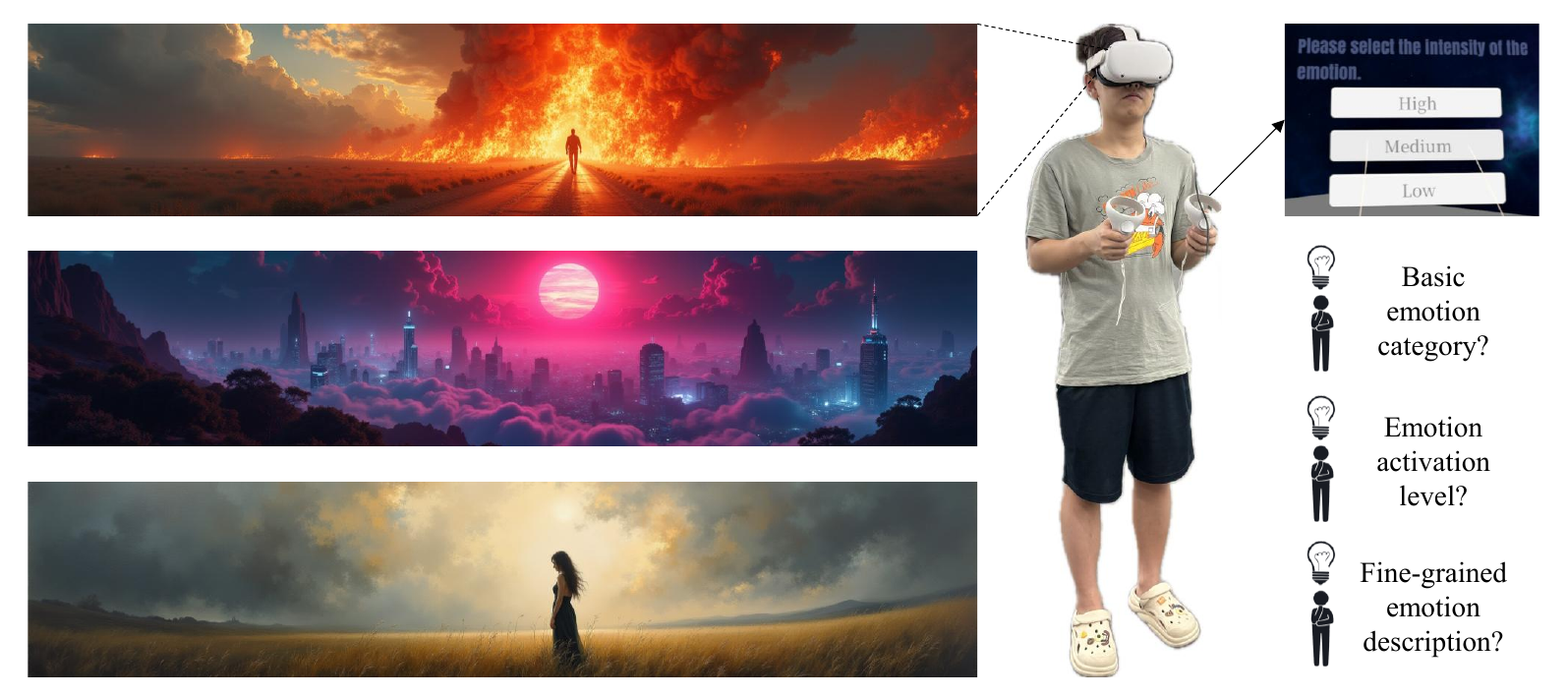}
\caption{\textbf{Experiment Setting.} Participants viewed and assessed emotional panoramas using a Meta Quest 2 headset and the in-VR response interface.}
\label{fig:setting}
}
\end{figure}

\rev{We used a Meta Quest 2 headset (FoV: 90°, frame rate: 90Hz, resolution: 1832$\times$1920 per eye)~\cite{metaquest2} for all experiments, with VR development in Unity 2022.3.6f1~\cite{unity2022}. The physical setup and in-headset assessment interface are shown in Fig.~\ref{fig:setting}. The headset supports natural head rotation and spatial inspection of the panoramas, while the desktop condition presents the same stimuli on a 27-inch monitor with mouse-based panning and zooming.}

\textbf{Participants.} 
\rev{The study protocol was approved by the Human Research Ethics Committee at HKUST(GZ) under HKUST(GZ)-HSP-2024-0073, and all participants provided informed consent before participation.}
We recruited 26 participants (P1--P26) aged 18--32 years, with 46.2\% female. Their occupations covered a wide range from interior designers to AI researchers. Participants reported varying VR usage frequency: 19.2\% use VR more than once a month, 23.1\% use VR from once a year to once a month, 42.3\% use VR less than once a year, and 15.4\% had never used VR before. All participants had normal (or corrected-to-normal) vision with no history or in-study experience of motion sickness. 

\rev{\textbf{Hypotheses.} We propose three hypotheses: (\textbf{H1}) Within VR, EmoSpace-generated panoramas are preferred over FLUX+IPR baselines for emotional expressiveness (method advantage); (\textbf{H2}) EmoSpace outputs are preferred as emotionally aligned and creatively suitable starting points in the creative design study; (\textbf{H3}) VR environments enhance subjective emotional experience compared to desktop (medium effect).}

\subsection{Quantitative Results}
\label{sec:user_study_results}

\begin{figure*}[htbp]
\rev{
\centering
\includegraphics[width=\linewidth]{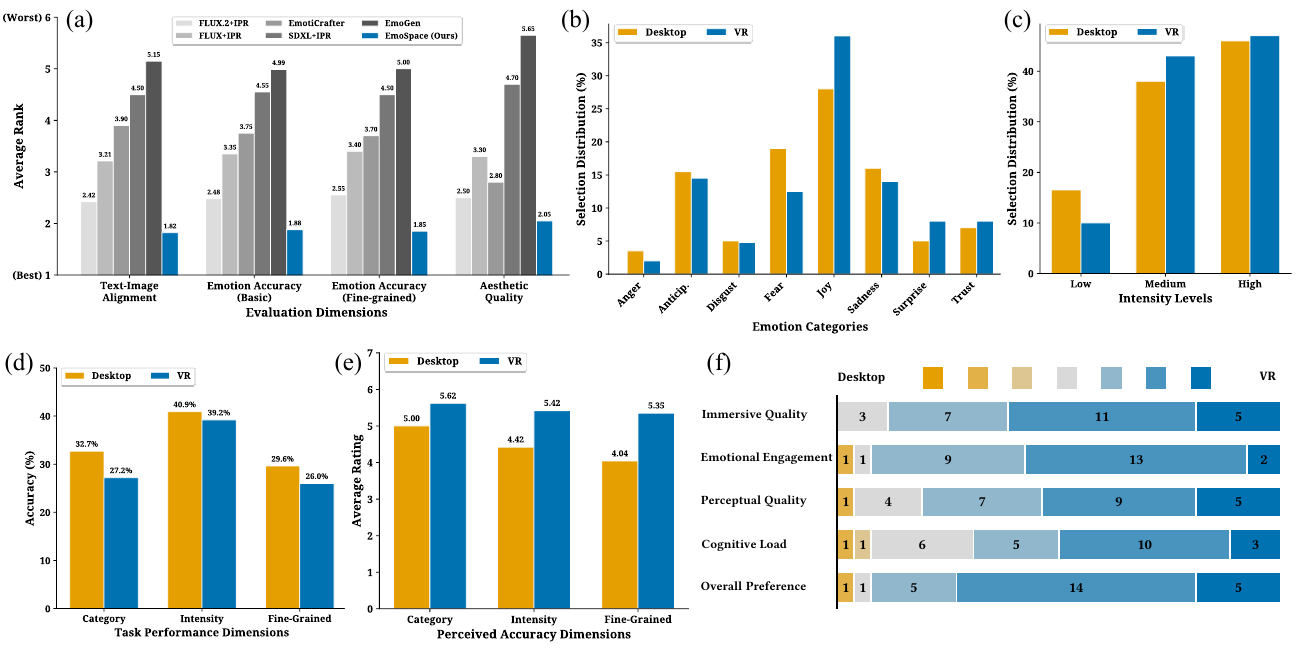}
\caption{\textbf{User Study Results.} (a) Human evaluation of \textit{EmoSpace} against other methods in text-image alignment, emotion accuracy (basic and fine-grained), and aesthetic quality. (b, c) VR Experiment 3: Differences in emotional category and intensity level perception between desktop and VR. (d, e) Participants' task performance and subjective perception in desktop and VR. (f) Post-study questionnaire results.}
\label{fig:task_performance}
}
\end{figure*}


\rev{\textbf{Experiment 1: Method-Specific Comparison.}
In the forced-choice tasks, participants selected EmoSpace over FLUX+IPR in 442 of 520 pairwise judgments for emotion accuracy (\textbf{T1}: \textbf{85.0\%}) and in 409 of 520 judgments for fine-grained discrimination (\textbf{T3}: \textbf{78.7\%}). Participants also rated EmoSpace significantly higher on emotional intensity (\textbf{T2}: $M_{\text{EmoSpace}} = 5.42$ vs.\ $M_{\text{FLUX}} = 4.18$, $t(25) = 6.23$, $p < 0.001$). These results support \textbf{H1}: the full EmoSpace framework yields method-associated differences in emotional expressiveness that remain perceptible when the display medium is held constant.}

\rev{\textbf{Experiment 2: Creative Design Preference Study.}
Participants selected EmoSpace-generated environments in 85 of 104 design choices (\textbf{81.7\%}). Because both candidates were rated separately, we used paired-samples $t$-tests with Holm correction across the three rating dimensions. EmoSpace received higher ratings than FLUX+IPR for emotional alignment ($M = 5.71$ vs.\ $4.32$, $t(25) = 7.45$), immersive potential ($M = 5.54$ vs.\ $4.61$, $t(25) = 5.12$), and creative suitability ($M = 5.68$ vs.\ $4.46$, $t(25) = 6.89$); all Holm-adjusted $p<0.001$. These results support \textbf{H2}: participants consistently preferred EmoSpace outputs as more emotionally aligned and more suitable starting points for the specified immersive design scenarios.}

\rev{\textbf{Experiment 3: Medium Comparison.}
Objective recognition accuracy is similar across VR and desktop conditions (Fig.~\ref{fig:task_performance}(d)): \textbf{T1} is 32.7\% versus 27.2\% (difference: 5.5 percentage points), \textbf{T2} is 40.9\% versus 39.2\% (1.7 points), and \textbf{T3} is 29.6\% versus 26.0\% (3.6 points). We report these participant-level accuracy summaries descriptively and do not infer a medium-related accuracy gain from the small differences. The marginal response distributions in Fig.~\ref{fig:task_performance}(b, c) additionally show more positive-category selections and fewer low-intensity selections in VR. Because aggregated marginal counts do not preserve the within-participant pairing, these distributional shifts are interpreted descriptively rather than through an independent-samples chi-square test.}

\rev{Fig.~\ref{fig:task_performance}(e) additionally shows higher participant-level perceived-accuracy ratings in VR across category, intensity, and fine-grained judgments. Because the corresponding objective accuracies in Fig.~\ref{fig:task_performance}(d) are comparable, this difference is interpreted as increased subjective confidence or perceptual engagement rather than improved recognition performance. As shown in Fig.~\ref{fig:task_performance}(f), Holm-corrected one-sample $t$-tests against the neutral midpoint (4.0) showed that participants rated VR higher for immersive quality ($M = 5.69$, $t(25) = 9.30$, $p < 0.001$), emotional engagement ($M = 5.50$, $t(25) = 7.73$, $p < 0.001$), perceptual quality ($M = 5.46$, $t(25) = 6.17$, $p < 0.001$), and overall preference ($M = 5.77$, $t(25) = 8.43$, $p < 0.001$). These subjective ratings support \textbf{H3}: VR environments enhance subjective emotional experience despite comparable objective performance, consistent with the ``circular interaction between presence and emotion''~\cite{riva2007affective}.}

\rev{Taken together, Experiments~1 and~2 show that the affective differences between \textit{EmoSpace} and the matched FLUX+IPR baseline remain perceptible in VR and influence design-oriented preferences. Experiment~3 shows that immersive presentation primarily changes subjective engagement and perceptual response patterns rather than objective recognition accuracy. The accompanying SDXL implementation further indicates that the prototype-guided conditioning concept can be adapted to a second backbone family through architecture-specific pathways.}

\subsection{Interview and Discussion}

In the semi-structured interview, we first asked participants about their overall emotional experience in the VR versus on the desktop. Users highlighted the enhanced immersive quality of VR environments (P1, P2, P6, P15, P17) and noted how the spatial environment affected their emotional experience (P9, P13, P20). The ability to observe details through dynamic viewpoints was particularly noted for its contribution to emotional engagement (P7, P11, P16, P18, P21). After this, they primarily focused on the following perspectives:

\rev{\textbf{Method Advantage and Creative Suitability.} Several participants noted that EmoSpace-generated environments felt ``\textit{more emotionally authentic}'' (P1, P4, P9, P15) and ``\textit{richer in emotional nuance}'' (P7, P11, P18). Several participants highlighted that EmoSpace panoramas contained ``\textit{subtle emotional cues that the baseline lacked}'' (P3, P8, P16), particularly for complex emotions like \textit{awe} and \textit{contentment}. This aligns with our quantitative finding that EmoSpace is selected for emotion accuracy in 85.0\% of the within-VR pairwise judgments. In the task-oriented study, participants appreciated the ``\textit{emotional precision}'' of EmoSpace environments (P2, P6, P13, P21), noting that the prototype-based approach produced images that ``\textit{already captured the emotional atmosphere I was aiming for}'' (P6, P11). Design-oriented participants (P1, P8, P14) emphasized that EmoSpace environments provided better starting points for further creative refinement.}

\textbf{Immersive Experience and Spatial Presence.} Users consistently reported that VR provided a more immersive experience compared to desktop environments, with participants describing feeling ``\textit{truly present in the scene}'' (P1, P2, P6, P15, P17, P19). The wide visual field and spatial enclosure created a sense of being ``\textit{wrapped in the environment}'' (P2, P9, P15, P20), which significantly enhanced emotional resonance. Users noted that VR allowed them to ``\textit{step into the picture rather than just viewing it}'' (P6, P11, P18), creating a fundamental shift from observer to participant. The ability to freely adjust viewing angles and distances was highlighted as a key factor in strengthening emotional connection (P7, P11, P16, P18, P21). Several participants emphasized the spatial volume and scale in VR amplified emotional impact, particularly for scenes with grandeur or intensity (P4, P9, P15, P19, P21).

\textbf{Emotional Intensity and Perception.} \rev{Participants reported stronger emotional intensity in VR than during desktop viewing across several categories (P1, P4, P9, P15, P18, P20, P22).} The enhanced detail perception in VR allowed users to notice facial expressions and environmental nuances that influenced their emotional judgments (P3, P8, P16, P17, P21). Moreover, users noted that the immersive environment facilitated deeper emotional engagement, with some describing how they could ``\textit{feel the emotions of characters in the scene}'' (P2, P6, P13, P21). The ability to view scenes from multiple perspectives was also credited with providing a more comprehensive understanding of emotional context (P7, P11, P18, P21).

\textbf{Cognitive Load and Detail Processing.} Contrary to our initial expectations, most users reported no increase in cognitive burden from the magnified visual presentation in VR (P1, P7, P10, P13, P20). Instead, participants described the detailed observation as enhancing rather than hindering their emotional understanding (P7, P8, P16, P17, P21). Users noted that while VR required more active exploration through head and hand movements, this was perceived as engaging rather than taxing (P5, P11, P19, P21). The ability to focus on specific details without external distractions was highlighted as a cognitive advantage of VR (P2, P15, P16, P20). Several participants mentioned that the immersive environment helped them concentrate better on emotional assessment tasks (P6, P9, P12, P18, P22).

\textbf{Interpretation of Perceptual Disconnect.} \rev{As summarized in Sec.~\ref{sec:user_study_results}, we observed a disconnect between objective accuracy and subjective perceived accuracy in \textbf{T2} and \textbf{T3}.} In the interview, some users also reported higher confidence and perceived accuracy when performing tasks in VR (P1, P2, P7, P9). We attribute this to the ``circular interaction between presence and emotion''~\cite{riva2007affective}: the immersive nature of VR amplifies the subtle emotional cues (P2, P7, P18, P21), making the intended affect feel more ``\textit{immediate},'' ``\textit{embodied}'' and ``\textit{authentic}'' (P1, P6, P9, P20), thus improving users' overall engagement and subjective confidence despite similar objective performance.

\rev{\textbf{Participant Feedback and Design Opportunities.} Participants highlighted several directions for future immersive systems, including lighter hardware for sustained use (P10, P14, P18, P24), dynamic and spatially interactive content (P6, P11, P16, P19, P21), richer emotion controls and custom descriptions (P7, P9, P15, P21, P23), and audio or multisensory feedback (P4, P15, P20, P25). They also identified potential applications in gaming, therapy, education, concentration support, and cultural preservation (P1, P8, P14, P18, P25).}

\section{\rev{Discussion}}

\rev{Beyond demonstrating the effectiveness of \textit{EmoSpace}, our experiments provide broader insights into the design of prototype-guided affective generation systems. This section discusses the implications of the proposed framework for immersive content generation, summarizes the lessons learned from our empirical studies, and highlights current limitations and future research directions.}

\rev{\textbf{Insights into Prototype-Guided Conditioning.}
Beyond the quantitative improvements reported in Secs.~IV and~V, our results provide several insights into the design of controllable affective generation systems. First, representing emotions through a hierarchical prototype bank provides a useful balance between semantic structure and expressive capacity. Unlike predefined categorical labels or low-dimensional affective spaces, the learned hierarchy is designed to organize fine-grained intra-category variations while preserving parent-emotion structure, and the dynamic-versus-static comparison indicates that input-conditioned adaptation contributes to controllable synthesis. Second, our experiments suggest that effective emotion control should be designed according to the conditioning pathways of the underlying diffusion architecture rather than relying on a universal injection strategy. The improvements associated with Prototype-Conditioned Steering suggest that architecture-aware conditioning can better exploit the available pathways of modern DiT backbones. Finally, immersive applications often require multiple control signals beyond emotion alone. Our results indicate that explicitly coordinating prototype-conditioned emotion modulation with external control conditions is associated with more stable emotional expression than applying the controls independently, providing practical guidance for designing unified multi-conditional generation frameworks.}

\rev{\textbf{Implications for Immersive Content Creation.}
Our VR studies further examine whether the affective differences associated with the framework remain perceptible and relevant in immersive settings. In Experiment~1, participants selected EmoSpace-generated content more often than the matched baseline within the same VR environment, indicating that the associated affective differences remain perceptible under immersive viewing conditions. Experiment~2 further shows that participants perceived EmoSpace-generated panoramas as more suitable starting points for iterative refinement and emotional storytelling in the creative design task. Together, these findings suggest that structured emotional control can provide useful candidate content for immersive design and motivate its integration into interactive authoring workflows.}

\rev{\textbf{Immersive Evaluation and Framework Generalization.}
The three experiments provide complementary evidence on generation quality and presentation effects. Experiments~1 and~2 show that the affective differences between \textit{EmoSpace} and the matched baseline remain perceptible in VR and influence design-oriented preferences, whereas Experiment~3 shows that immersive presentation primarily enhances subjective emotional engagement, while objective recognition accuracy remains broadly comparable across conditions. Furthermore, implementations on FLUX.1-dev (DiT) and SDXL (U-Net) show that prototype-guided conditioning can be instantiated through backbone-specific pathways in the two evaluated architecture families.}

\rev{\textbf{Limitations and Future Work.} Despite encouraging results, several limitations remain. First, the current framework focuses on static image generation; extending prototype-guided conditioning to video generation will require explicitly modeling temporal consistency of emotional expression across frames. Second, although our hierarchical prototype space is designed to represent fine-grained sub-emotion variation, it does not explicitly model emotion intensity or mixed emotional states, which commonly occur in real-world affective experiences. Third, the creative design study is selection-based and does not evaluate end-to-end authoring efficiency or learnability; future work should examine interactive VR workflows with iterative generation and editing. Fourth, no widely adopted benchmark currently covers fine-grained emotion-conditioned panorama, style, and VR generation jointly. We therefore evaluate the framework under a fixed and reproducible protocol and will release the prompt suite, random seeds, generated samples, and evaluation scripts. Finally, although our framework is validated on both DiT and UNet backbones, more extensive evaluations on emerging foundation models and standardized benchmarks for affective image generation would further strengthen reproducibility and cross-model comparisons. Future work will explore personalized emotion modeling, spatio-temporal emotional narratives, and multi-sensory affective interaction to support more intelligent and emotionally adaptive immersive experiences.}


\section{Conclusion}

\rev{This paper presented \textit{EmoSpace}, a unified prototype-guided conditioning framework for immersive affective content generation built upon the FLUX DiT architecture. By transforming a hierarchical emotion representation into structured conditioning signals, our framework supports fine-grained affective control through architecture-aware conditioning together with panoramic, stylized, and multi-conditional generation. Across the reported evaluations, EmoSpace obtains the highest values on the emotion-focused automatic metrics, preserves aesthetic quality comparable to the strongest baseline, and receives favorable controlled human judgments. The VR studies further separate method-associated preferences from the influence of immersive presentation on emotional perception.}

\rev{More broadly, our work demonstrates the potential of prototype-guided conditioning as a promising paradigm for controllable affective generation with modern diffusion transformers. We hope it inspires future research on emotionally intelligent generative models for immersive media, including personalized affective generation, spatio-temporal emotional narratives, and multi-sensory interaction.}





\bibliographystyle{IEEEtran}
\bibliography{template}


 





\vfill

\end{document}